\documentclass[sigconf]{acmart}
\usepackage{graphicx}
\usepackage{amsthm,amsmath,amsfonts}
\usepackage{subfigure}
\usepackage{color}
\usepackage{bm}
\usepackage{multirow}
\usepackage{array}
\usepackage{algorithm} 
\usepackage{algorithmicx}
\usepackage{algpseudocode}

\def\bA{\textbf{A}}

\def\bH{\textbf{H}}
\def\bW{\textbf{W}}

\def\bh{\textbf{h}}

\def\bal{\bm{\alpha}}
\newcolumntype{L}[1]{>{\raggedright\let\newline\\\arraybackslash\hspace{0pt}}m{#1}}
\newcolumntype{C}[1]{>{\centering\let\newline  \\\arraybackslash\hspace{0pt}}m{#1}}
\newcolumntype{R}[1]{>{\raggedleft\let\newline \\\arraybackslash\hspace{0pt}}m{#1}}

 \usepackage{multirow}
 \usepackage[normalem]{ulem}
\useunder{\uline}{\ul}{}
\DeclareMathOperator*{\argmin}{arg\,min}
\usepackage{colortbl}

\newtheorem{prop}{Proposition}

\copyrightyear{2023}
\acmYear{2023}
\setcopyright{acmlicensed}\acmConference[WWW '23]{Proceedings of the ACM Web Conference 2023}{May 1--5, 2023}{Austin, TX, USA} \acmBooktitle{Proceedings of the ACM Web Conference 2023 (WWW '23), May 1--5, 2023, Austin, TX, USA}
\acmPrice{15.00}
\acmDOI{10.1145/3543507.3583486} \acmISBN{978-1-4503-9416-1/23/04}

\begin{document}

\title{Search to Capture Long-range Dependency with Stacking GNNs for Graph Classification}


\author{Lanning Wei$^\dagger$}
\affiliation{%
	\institution{Institute of Computing Technology, \\ Chinese Academy of Sciences \\ University of Chinese Academy of Sciences}
	\city{Beijing}
	\country{China}
}
	\email{weilanning18z@ict.ac.cn}

\author{Zhiqiang He$^\dagger$}
\affiliation{%
	\institution{Institute of Computing Technology, \\ Chinese Academy of Sciences \\ Lenovo}
	\city{Beijing}
	\country{China}
}
\email{hezq@lenovo.com}

\author{Huan Zhao$^\ast$}
\affiliation{%
	\institution{4Paradigm. Inc}
	\city{Beijing}
	\country{China}
}
\email{zhaohuan@4paradigm.com}
\author{Quanming Yao}
\affiliation{%
	\institution{Department of Electronic Engineering, \\ Tsinghua University}
	\city{Beijing}
	\country{China}
}
\email{qyaoaa@tsinghua.edu.cn}
	
\renewcommand{\shortauthors}{Lanning Wei, Zhiqiang He, Huan Zhao, Quanming Yao.}

\begin{abstract}
	In recent years, Graph Neural Networks (GNNs) have been popular in the graph classification task.
	Currently, shallow GNNs are more common due to the well-known over-smoothing problem facing deeper GNNs. However, they are sub-optimal without utilizing the information from distant nodes, i.e., \textit{the long-range dependencies}.
	The mainstream methods in the graph classification task can extract the long-range dependencies either by designing the pooling operations or incorporating the higher-order neighbors, while they have evident drawbacks by modifying the original graph structure, which may result in information loss in graph structure learning.
	In this paper, by justifying the smaller influence of the over-smoothing problem in the graph classification task, we evoke the importance of stacking-based GNNs and then employ them to capture the long-range dependencies without modifying the original graph structure.
	To achieve this,  two design needs are given for stacking-based GNNs, i.e., sufficient model depth and adaptive skip-connection schemes.
	By transforming the two design needs into designing data-specific inter-layer connections, we propose a novel approach with the help of neural architecture search (NAS), which is dubbed LRGNN (Long-Range Graph Neural Networks).
	Extensive experiments on five datasets show that the proposed LRGNN can achieve the best performance,
	and obtained data-specific GNNs with different depth and skip-connection schemes, which can better capture the long-range dependencies.
	\footnote{
		$\dagger$: Both authors contributed equally to this paper. $\ast$: Corresponding author. The implementation of LRGNN is available at: \url{https://github.com/LARS-research/LRGNN}.}
\end{abstract}



\maketitle

\section{Introduction}
\label{sec-intro}

In recent years, Graph Neural Networks (GNNs) have been the state-of-the-art (SOTA) method on graph classification~\cite{zhang2018end,wang2021curgraph}, a popular task which can be applied into various domains, e.g., chemistry~\cite{gilmer2017neural}, bioinformatics~\cite{ying2018hierarchical,wang2022graph}, text~\cite{zhang2020every} and social networks~\cite{xu2018powerful,wang2021mixup,yoo2022model}.
In general, they encode the graph structure to learn node embeddings with the help of the message-passing scheme~\cite{gilmer2017neural}, i.e.,  aggregating messages from the connected nodes. Then, the graph representation vector can be generated with one readout operation, e.g., take the mean or summation of all node embeddings.
In the literature,  two- or three-layer GNNs are widely used since the performance decreases as the network goes deeper~\cite{li2018deeper}.  
However, it limits the ability of GNNs to capture the \textit{long-range dependencies}, 
i.e., incorporate information from long-distant neighbors~\cite{lukovnikov2021improving}, which may be useful in the graph classification task~\cite{jain2021representing,dwivedi2022long}.

For the graph classification task, the majority of methods focus on designing the pooling operations to extract the hierarchical information in the graph~\cite{lee2019self,ying2018hierarchical}.
We review the literature
from the perspective of long-range dependencies, and then observe that not all pooling operations are helpful in extracting the long-range dependencies.
The grouping-based pooling operations~\cite{ying2018hierarchical,bianchi2020spectral,yuan2019structpool} group nodes into clusters and then re-design the edges between these clusters.
These newly added edges may shorten the distance between node pairs, and then faster the feature propagation in the graph. 
On the contrary, the selection-based ones~\cite{lee2019self,gao2019graph} construct the coarse graph by removing the low-ranked nodes and edges connected with them, 
which means the distance between nodes will not be shortened and are not helpful in incorporating the long-distant neighbors.
Apart from pooling operations, several methods tend to incorporate the higher-order neighbors directly.
For example, 
GraphTrans~\cite{jain2021representing} adopts the transformer module which allows the interactions between all node pairs. Similarly, \cite{pham2017graph} provides one virtual node which connected with all nodes, and \cite{alon2020bottleneck} designs one fully adjacent (FA) aggregation layer in which all nodes are connected directly.  

Despite the popularity of these methods in graph classification, they are deficient when capturing long-range dependencies on graphs. The aforementioned methods update the graph structure and may result in information loss in graph structure learning, either by generating the coarse graph or connecting to higher-order neighbors. 
Considering this, we revisit the stacking-based GNNs, which is a direct approach to obtaining the long-range dependencies while keeping the graph structures unchanged.
%
In general, scaling the model depth is the most common way to improve the model performance~\cite{tan2019efficientnet,krizhevsky2017imagenet,simonyan2014very,he2016deep}.
and deeper stacking-based GNNs further bring the larger receptive field to incorporate the longer-distant neighbors.
Nevertheless, the major concern to utilize the deeper GNNs is the over-smoothing problem~\cite{li2018deeper,xu2018representation},
i.e., the connected nodes tend to have similar features as the network goes deeper, which results in a performance drop in the node classification task. 
Yet, in this paper, we justify the smaller influence of this problem on the graph classification task compared with the node classification task both in theoretical analysis and experimental results. Therefore, designing stacking-based GNNs is indeed a feasible solution to capture the long-range dependencies.

Motivated by this, we propose a novel method LRGNN (\underline{L}ong-\underline{R}ange \underline{G}raph \underline{N}eural \underline{N}etworks), and employ the stacking-based GNNs to capture the long-range dependencies in graphs. 
There are two aspects that will affect the utilization of the long-distant neighbors:
(a) sufficient model depth is required to incorporate the longer-distant neighbors; (b) adaptive skip-connection schemes are required considering the information mixing from different ranges of neighbors. 
The former is correlated with the inter-layer connections between the consecutive layers while the latter one related to the inconsecutive layers. 
Therefore, designing effective stacking-based GNNs to capture the long-range dependencies can be achieved by designing the inter-layer connections adaptively.
We adopt the widely used neural architecture search (NAS) methods to achieve this.
%

To be specific, we provide one framework to design the inter-layer connections, which contains a set of learnable connections in the directed acyclic graphs (DAG). Two candidate choices are provided to represent the ``used'' and ``unused'' states in learnable connections, and then designing the inter-layer connections is transformed into deciding which choice should be adopted in each learnable connection. 
Then, the search space of LRGNN can be constructed based on the combinations of those connection choices, while the search space size is growing exponentially as the network goes deeper.
Considering the search efficiency, we further provide one cell-based framework that only enables the connection designs in each cell, on top of which two variants LRGNN-Diverse and LRGNN-Repeat can be constructed.
The differentiable search algorithm is adopted in this paper to enable the adaptive architecture design.
In the experiments, we first evaluate the rationality of LRGNN by showing the higher performance achieved with sufficient model depth and various skip-connection schemes.
Then, the extensive experimental results demonstrate that the proposed LRGNN can achieve the SOTA performance by designing GNNs with the inter-layer connections adaptively, on top of which the effectiveness of the proposed method can be verified.

To summarize, our contributions are as follows:

\begin{itemize}
\item We evoke the importance of stacking-based GNNs in extracting the long-range dependencies, and verify the smaller influence of the over-smoothing problem on the graph classification, on top of which two design needs are provided for designing stacking-based GNNs , i.e., the sufficient GNN depth and adaptive skip-connection schemes.

\item To meet these two design needs, we firstly unify them into designing the inter-layer connections in stacking-based GNNs, and then achieve this with the help of NAS.

\item We conduct extensive experiments on five datasets from different domains, and the proposed LRGNN achieves the SOTA performance by designing GNNs with sufficient model depth and adaptive skip-connection schemes.
\end{itemize}

\noindent\textbf{Notations.}
We represent a graph as $\mathcal{G} =(\mathcal{V}, \mathcal{E}) $,where $\mathcal{V}$ and $\mathcal{E}$ represent the node and edge sets. $\textbf{A} \in \mathbb{R}^{|\mathcal{V}| \times |\mathcal{V}|}$ is the adjacency matrix of this graph where $|\mathcal{V}|$ is the node number. $\mathcal{N}(u)$ is the neighbors of node $u$. $\bH \in \mathbb{R}^{|\mathcal{V}| \times d}$ is the node feature matrix and $d$ is the feature dimension, and $\bh_u$ is the feature representation of node $u$.

\section{Related Work: Graph neural network~(GNN)}
\label{sec-pre-mpnn}

GNNs have advantages in encoding the graph structure information with the help of the message-passing scheme~\cite{gilmer2017neural}, 
i.e., aggregating the messages from connected nodes. It can be represented as $\bh_v = \bW(\bh_v, \text{AGGR}\{\bh_u, u \in \mathcal{N}(v)\})$, 
where $\bW$ is the learnable parameter, \text{AGGR} is the aggregation function used in this aggregation operation.  Diverse aggregation operations are proposed and widely used in graph representation learning~\cite{kipf2016semi,hamilton2017inductive,velivckovic2017graph,chen2020simple}.
Based on these aggregation operations, one GNN can be constructed by stacking these aggregation operations. 

\subsection{Long-range Dependency Extraction in Graph Classification }
\label{sec-related-longrange}
In the graph classification task, existing literature can be grouped into three categories according to their methods in extracting the long-range dependencies, i.e., designing pooling operations, incorporating the higher-order neighbors, and stacking GNN layers.

\subsubsection{Designing pooling operations}
\label{sec-related-pooling}
Pooling operations are widely used in the graph classification task, and they aim to generate one coarse graph to extract the hierarchical information. 
They can be classified into two groups, i.e., the selection-based ones and the grouping-based ones.
The selection-based methods~\cite{lee2019self,gao2019graph} focus on evaluating the node importance. 
The top-ranked nodes are preserved, and then they construct the coarse graph by dropping the edges connected with the un-selected nodes.
On the contrary, the grouping-based methods~\cite{ying2018hierarchical,bianchi2020spectral,yuan2019structpool} aim to group nodes into several clusters based on their similarities. 
They first design one assignment matrix, on top of which the cluster features and new edges are constructed.

Despite the success of these methods in the graph classification task, not all pooling operations help obtain the long-range dependencies in the graph.
The selection-based operations drop the edges connected with the un-selected nodes, and then the shortest path between node pairs will not decrease. Therefore, these pooling operations cannot faster the feature propagation in the graph.
The grouping-based pooling operations reconstruct the edges in the coarse graph, and these edges may shorten the distance between the node pairs.

\subsubsection{Incorporating higher-order neighbors}
\label{sec-related-higherorder}
The general aggregation operation only propagates messages from the connected neighbors.
By connecting nodes with higher-order neighbors, 
the long-range dependencies can be obtained with fewer aggregation operations.
In the graph classification task, \cite{pham2017graph} provides the virtual node which is connected with the other nodes, on top of which the distance for each node pair is less than two. 
\cite{alon2020bottleneck} provides a fully-adjacent layer at the end of GNNs, in which every pair of nodes are connected with an edge.
Transformer modules~\cite{vaswani2017attention} are designed to communicate with other nodes. Existing methods designed diverse positional encoders to learn graph structures.
For example, GraphTrans~\cite{jain2021representing} uses the stacked GNNs to encode the graph structure.
Graphormer~\cite{ying2021transformers} designs three encoders to embed the node centrality, node pairs, and edge features separately, and then applied the Transformer on these node sets. 

\subsubsection{Stacking GNN layers}
\label{sec-related-stacking}

Apart from designing specific operations to obtain the long-range dependencies, stacking more GNN layers can incorporate the message from longer-distant neighbors. To be specific, each aggregation operation aggregates messages from the connected neighbors, on top of which the receptive field of each node can be expanded one hop away. Therefore, for one $k$-layer stacking GNN, each node can incorporate the messages from its $k$-hop neighbors. 
Therefore, to extract the long-range dependencies which may be important in the graph classification task, one GNN can stack more aggregation operations.
In the graph classification task, GIN~\cite{xu2018powerful} aims to design power aggregation operations, and then five layers are applied in GNN. DGCNN~\cite{zhang2018end} stacks three graph convolution layers, and then ranks nodes based on the node features, on top of which the graph representations can be generated with those top-ranked ones.

\subsection{Graph Neural Architecture Search}
\label{sec-related-gnas}
Researchers tried to design GNN architectures by neural architecture search (NAS) automatically~\cite{wang2022automated}.
 The majority of these methods focus on designing the aggregation layers in GNNs, e.g., GraphNAS~\cite{gao2019graphnas} and GraphGym~\cite{you2020design} provide diverse dimensions and candidate operations to design the GNN layers. Besides, SANE~\cite{zhao2021search}, SNAG~\cite{zhao2020simplifying}, F2GNN~\cite{wei2022designing} and AutoGraph~\cite{li2020autograph} design the skip-connections based on the stacked aggregation operations.
Policy-GNN~\cite{lai2020policy} and NWGNN~\cite{wang2022nwgnn} aim to design the GNN depth.
Apart from these methods designed for the node classification task, NAS-GCN~\cite{jiang2020graph} learns adaptive global pooling functions additionally, and PAS~\cite{wei2021pooling} is proposed to design global and hierarchical pooling methods adaptively for the graph classification task.
Despite the success of these methods, they are usually shallow, e.g., use two- or three-layer GNNs in general.
DeepGNAS~\cite{feng2021search} designs deep architectures in the block and architecture stages for the node classification task. 
As to the search algorithm, 
differentiable search algorithms are preferred in recent years~\cite{liu2018darts} considering the search efficiency. The discrete search space is relaxed into continuous with one relaxation function~\cite{liu2018darts,xie2018snas}, on top of which the gradient descent can be applied.

More graph neural architecture search methods can be found in~\cite{zhang2021automated,wang2022automated,wang2021autogel,wang2022profiling}. Compared with existing methods which use shallow GNNs or only design deep GNNs in the node classification task, LRGNN evokes the importance of deep stacking-based GNNs in the graph classification task, and the rationality has been justified in Section~\ref{sec-method-twoneeds}. 

\section{Method}
In this section, we first show the feasibility of employing the stacking-based GNNs to extract the long-range dependencies, on top of which two design needs for these GNNs are provided. Then, we will introduce the proposed method LRGNN.
 
\subsection{The Feasibility of Stacking-based GNNs}
\label{sec-method-twoneeds}
For the graph classification task, one GNN aims to learn one representation vector for each input graph to predict the graph label, and it is related to the interactions between long distance pairs~\cite{jain2021representing}, e.g., counting local substructures~\cite{dwivedi2022long}. Therefore, how effectively embracing long-range dependencies is the key factor in designing neural networks. 
\label{sec:method}
%
%
However, existing methods mentioned in Section~\ref{sec-related-pooling} and \ref{sec-related-higherorder} are deficient in capturing long-range dependencies. These methods update the graph structures either by generating the coarse graph or connecting to higher-order neighbors, which may result in insufficient discriminability, i.e., the discriminative graph structures may become indistinguishable anymore. For example, the pooling operations may generate the same coarse graph based on two distinguishable graphs, and the instances are provided in Appendix~\ref{sec-appendix-pooling-illustrations}.

Considering the deficiency of these methods, we turn to the stacking-based GNNs and search for better solutions to capture long-range dependencies.
In general, scaling the model depth is the common way to improve the model performance in computation vision~\cite{tan2019efficientnet,he2016deep}, and deeper GNNs can enlarge the receptive field which enables the extraction of longer-range dependencies.
Although, the over-smoothing problem is hard to evade in deep GNNs~\cite{li2018deeper,xu2018representation}, i.e., the connected nodes will have similar representations as the network becomes deeper, which will result in a performance drop on the node classification task. This problem hinders the development of deeper GNNs, and two- or three-layer shallow GNNs are widely used.
However, as shown in proposition~\ref{prop-smooth}, we theoretically justify that the over-smoothing problem has smaller influence on the graph classification task compared with the node classification task. The proof is provided in Appendix~\ref{sec-appendix-prop2}. 
Therefore, it is one potential solution to obtaining the long-range dependencies by stacking sufficient GNN layers.
In this paper, we extract the long-range dependencies by stacking sufficient GNN layers as mentioned in Section~\ref{sec-related-stacking}.
Compared with the aforementioned methods, it can preserve the graph structure information in the computational graph without modifying the graph structure.

\begin{figure}[ht]
	\centering
	\includegraphics[width=0.95\linewidth]{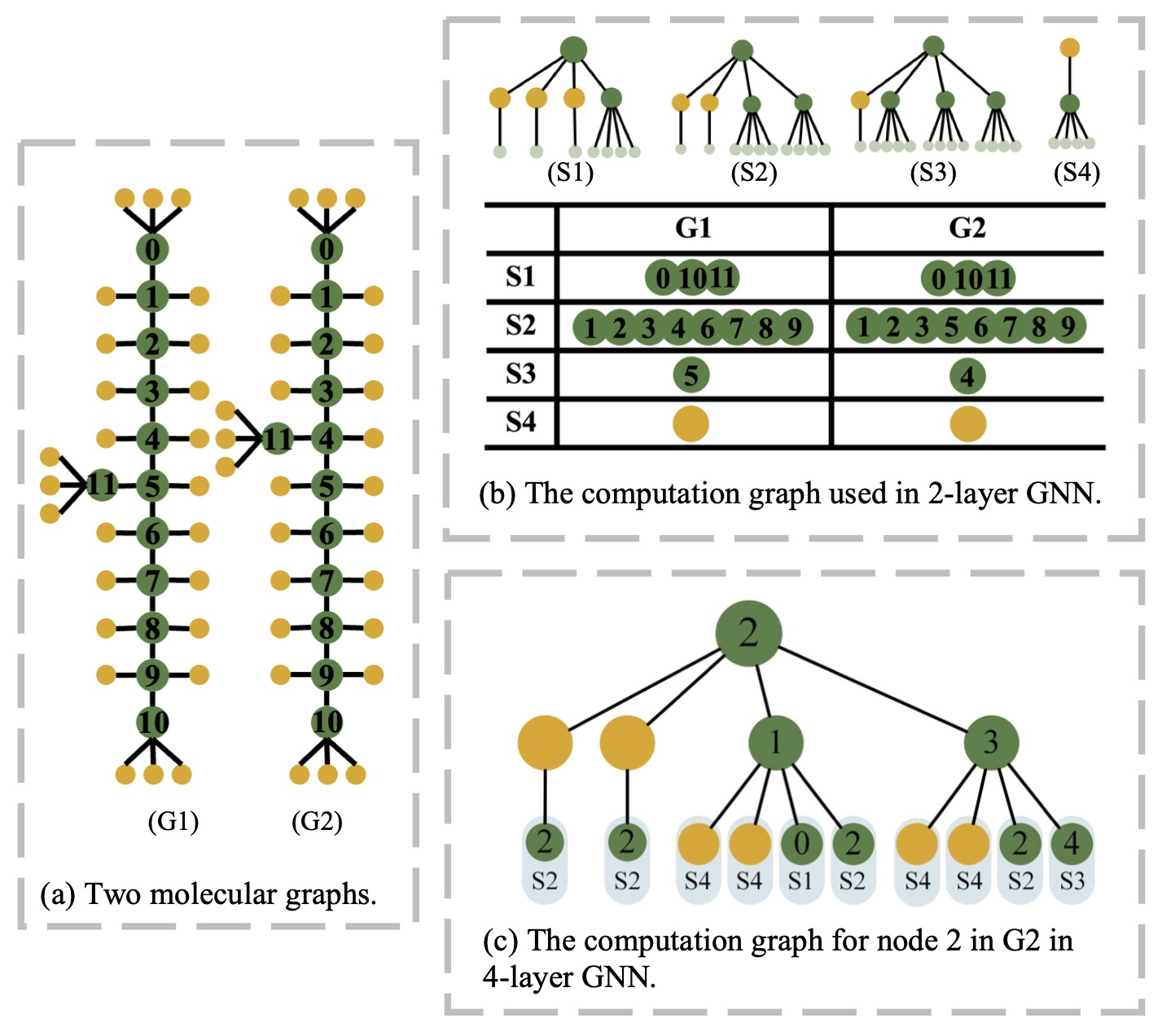}
	\vspace{-10pt}
	\caption{(a): The illustrations of two molecule graphs ($C_{12}H_{26}$) where the green node is \textbf{carbon} and the orange node is \textbf{hydrogen}. 
		(b): We visualize the computation graph used in the 2-layer GNNs for each node in the table. 
		Two graphs have the same subtree distribution and lead to the same graph representation.
		(c): Especially, we visualize the computation graph for node 2 in $\mathcal{G}_2$ used in the 4-layer GNN. It contains $S1$ and $S3$ simultaneously, which is unique in $\mathcal{G}_2$. }
	\label{fig-long-range-example}
	\vspace{-10pt}
\end{figure}

\begin{prop}
	Let $\mathcal{G}_1$ and $\mathcal{G}_2$ are two different graphs that satisfy: (a) have no bipartite components, and (b) at least one of the eigenvectors or eigenvalues are different.
	Then, for any learnable parameter $\bW$, $(I-L_1^{sym})^{\infty}\bW \neq (I-L_2^{sym})^{\infty}\bW$, where $L_1^{sym}$ and  $L_2^{sym}$ are the symmetric Laplacian matrices of $\mathcal{G}_1$ and $\mathcal{G}_2$, respectively.
	\label{prop-smooth}
\end{prop}



\subsection{Design Needs for Stacking-based GNNs}
When using the stacking-based GNNs to capture the long-range dependencies, two aspects affect the utilization of long-distant neighbors, i.e., the GNN depth and the skip-connection schemes. The former reflects the longest distance one node can access, and the latter is related to 
the information mixing from different ranges of neighbors.
Therefore, to effectively capture the long-range dependencies, these two design needs should be considered when designing the stacking GNNs in this paper.

\subsubsection{Depth needs to be sufficient.}
\label{sec-sufficient-depth}

By stacking more aggregation operations, deeper GNNs can incorporate the interactions from more distant neighbors which affect the prediction results.  We theoretically justify the discriminative power, i.e., how well GNNs with different depths can distinguish the non-isomorphic graphs, in Proposition~\ref{prop-long-range}. 
The proof is proved in Appendix~\ref{sec-appendix-prop1}. Use the cases shown in Figure~\ref{fig-long-range-example} (a) as an example,
the key difference between these two graphs is the graph structures within nodes 0 and 11, and these two graphs can be distinguished by the 1-WL test in the $4$-th iteration. 
As shown in Figure~\ref{fig-long-range-example} (b), two-layer GNN cannot distinguish two graphs since the same computation graphs are obtained in two molecular graphs, and the same graph representation vector will be obtained based on this. 
However, as shown in Figure~\ref{fig-long-range-example} (c), the visualized computation graph of node $2$ in $\mathcal{G}_2$ contains $S1$ and $S3$ simultaneously, while it cannot be achieved in $\mathcal{G}_1$. On top of this, a four-layer GNN can distinguish these two graphs.
Combined with the 1-WL test, deeper GNNs are more expressive than shallower ones, while two- or three-layer GNNs are widely used in the graph classification task. Therefore, it is necessary to enlarge the model depth when capturing the long-range dependencies in the graph classification.

\noindent\begin{prop}
	For any two graphs $\mathcal{G}_1$ and $\mathcal{G}_2$ which are non-isomorphic and can be distinguished by the first-order Weisfeiler-Lehman test in $k$-th iteration, one aggregation-based $L$-layer GNN $\mathcal{A}_L : \mathcal{G}\rightarrow\mathbb{R}^d$ can come up the following conclusions:
	\begin{align}
		& \mathcal{A}_L(\mathcal{G}_1) \neq \mathcal{A}_L(\mathcal{G}_2), L \geq k,  \nonumber \\
		& \mathcal{A}_l(\mathcal{G}_1) = \mathcal{A}_l(\mathcal{G}_2), \forall \, 0 \leq l< k. \nonumber
	\end{align}
	\label{prop-long-range}
	\vspace{-10pt}
\end{prop}

\begin{figure}[ht]
	\centering
	\includegraphics[width=0.85\linewidth]{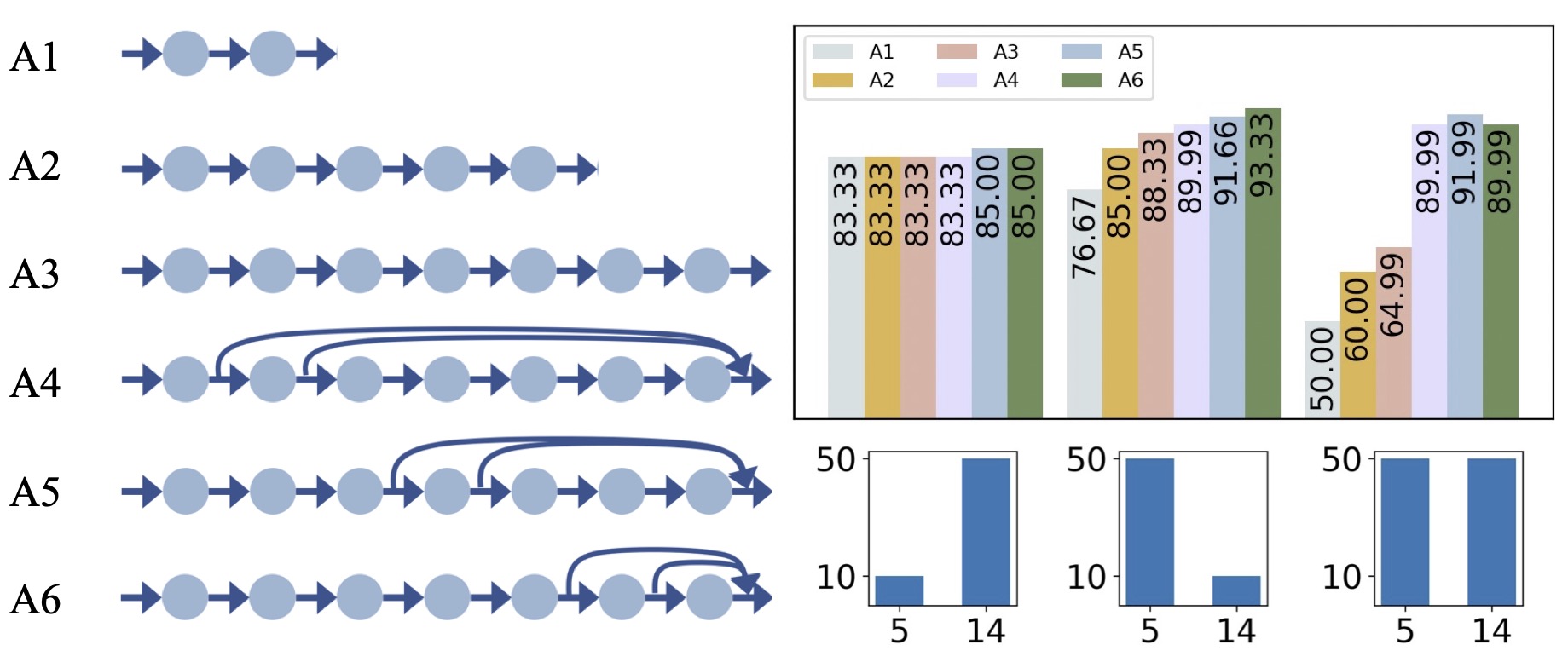}
	\vspace{-10pt}
	\caption{The performance comparisons with GNNs in different model depth and skip-connection schemes. Each dataset has a different number of graphs in diameter of $5$ and $14$. Graph label is determined by its diameter. The details are provided in Appendix~\ref{sec-appendix-adaptive-exp}.}
	\vspace{-10pt}
	\label{fig-diameter-connection}
\end{figure}

 \subsubsection{Skip-connection needs to be adaptive.}
 \label{sec-adaptive-connection}
Based on proposition~\ref{prop-long-range}, different graph pairs may have different distinguishable iteration $k$ in the 1-WL test. Therefore, it is unreasonable to incorporate the distant neighbors with only one variety, i.e., only from the $k$-hop away, when facing those graphs in the datasets. 
In stacking-based GNNs,  each node can increasingly expand the receptive field as the network goes deeper,  and then designing the skip-connection schemes in GNN can make up for this deficiency since the information extracted from different ranges of neighbors can be mixed based on the skip-connections, on top of which the extracted information can be enriched.

As shown in Figure~\ref{fig-diameter-connection}, we design six GNNs with different model depths and skip-connection schemes, and then evaluate them on three datasets that have different graph diameter distributions. We observe that: (a) deeper GNNs achieve higher performance in general, which demonstrates the importance of sufficient large layer numbers in the graph classification task; (b) different skip-connection schemes result in diverse performances in each dataset, and they also have different ranks. Therefore, it is a necessity to design the skip-connections data-specifically~\cite{zhao2020simplifying,zhao2021search}.

\subsection{Designing Inter-layer Connections in Stacking-based GNNs}
\label{sec-design-space}
When using the stacking-based GNNs to extract the long-range dependencies, the model depth and the skip-connection schemes should be considered. These two designing needs can be unified by designing inter-layer connections, which correspond to the inter-layer connections between the consecutive and inconsecutive layers, respectively.
To effectively extract the long-range dependencies, we adopt NAS to design the inter-layer connections in GNNs, on top of which the GNN depth and skip-connection schemes will be obtained adaptively. 

\begin{figure}[ht]
	\centering
	\vspace{-10pt}
	\includegraphics[width=0.95\linewidth]{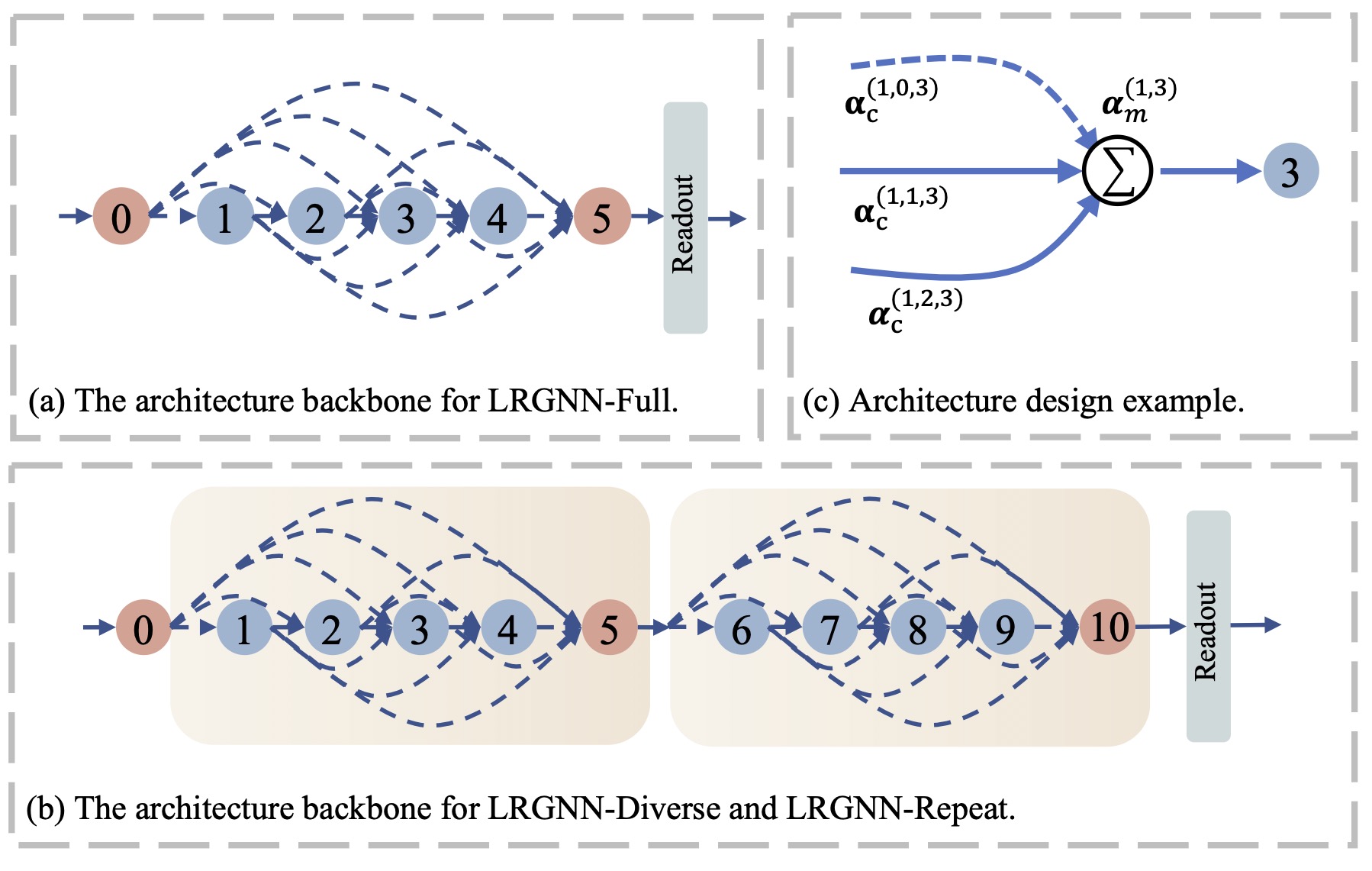}
	\vspace{-10pt}
	\caption{(a) GNN framework used to design the inter-layer connections. 
		(b) Cell-based GNN architecture. Each cell contains $B/C$ ($B=8 $ and $ C=2$ for example) aggregation operations and a post-processing operation. 
		(c) One architecture design example. 
		For aggregation operation 3, the connection from operation $0$ is unused, 
		and two inputs from operation $1$ and $2$ are merged with operation \texttt{SUM}, on top of which the aggregation operation $3$ can be operated.}
	\label{fig-cell-framework}
	\vspace{-10pt}
\end{figure}

\subsubsection{Architecture framework} 
As shown in Figure~\ref{fig-cell-framework} (a), we provide a unified framework to design the inter-layer connections in GNNs, 
which is constructed with an ordered sequence of $B$ aggregation operations ($B=4$ and GCN~\cite{kipf2016semi} operation for example).
Motivated by GraphGym~\cite{you2020design}, we provide one pre-processing operation $0$ and one post-processing operation $B+1$, each of which is one two-layer MLP (Multilayer Perceptron) to support the integration in the following procedures. 
At the end of this framework, one readout operation is provided to learn the graph representations.
By deciding whether to use these inter-layer connections, i.e., the dashed lines shown in the framework, we can obtain GNNs with different depths and skip-connections to extract the long-range dependencies.


\subsubsection{Designing search space based on the framework.}
We provide a set of candidates to construct the search space.
As shown in Figure~\ref{fig-cell-framework} (a), the connections among different operations (with dashed lines) are learnable, and each of them is an ``on-off switch'' when utilizing features, i.e., only ``used'' and ``unused'' states existed. Therefore, we provide two candidate choices to represent these two states on each learnable connection in the framework, i.e., $\mathcal{O}_c=\{$\texttt{ON}, \texttt{OFF}$\}$. They can be represented as $o(\bH)=\bH$ and $o(\bH)=\textbf{0}$, respectively.
For aggregation operations and the post-processing operation, more than one connection may be used.
A straightforward manner is to add up these selected features. To further improve the model expressiveness, we provide one merge operation to incorporate these selected features. 
Based on the literature, we provide six candidates to merge them with the concatenation, LSTM cell, attention mechanism, summation, average and maximum. Then the merge operations are denoted as  $\mathcal{O}_m=\{$\texttt{CONCAT}, \texttt{LSTM}, \texttt{ATT}, \texttt{SUM}, \texttt{MEAN}, \texttt{MAX} $\}$. 
For readout operation, we provide seven candidates to obtain the graph representation vector and it can be denoted as  $\mathcal{O}_r=\{$\texttt{GMEAN}, \texttt{GMAX}, \texttt{GSUM}, \texttt{GSORT},  \texttt{GATT}, \texttt{SET2SET }, \texttt{MEMA}$\}$. These candidates represent:
three simple global mean, max and sum functions; four readout operations derived from \cite{zhang2018end}, \cite{li2015gated}, \cite{vinyals2015order} and \cite{lee2019self}, respectively.

As shown in Figure~\ref{fig-cell-framework} (a), the operations used in each learnable connection are designed independently, and then different depth and skip-connection schemes will be obtained in GNNs. Meanwhile, designing the merge and readout operations can further improve the expressiveness of GNNs ~\cite{xu2018powerful}. This variant is denoted as LRGNN-Full in this paper.

As mentioned in Section~\ref{sec-sufficient-depth}, the GNN depth should be sufficiently large, which demonstrates the large number of aggregation operations in the framework. Then, the search space size is growing exponentially along with the increasing number of aggregation operations. 
Considering the search efficiency, we adopt the cells which are widely used in NAS~\cite{liu2018darts,pham2018efficient} and provide one cell-based framework as shown in Figure~\ref{fig-cell-framework} (b). 
Each cell can be treated as one basic unit to design the stacking-based GNNs.
$C$ cells are provided and then these $B$ aggregation operations are assigned to the cells equally. 
The operations in each cell can be learned independently, and we will obtain different cells with diverse connection schemes and merge operations in the searched GNN, which is dubbed LRGNN-Diverse. On the contrary, we can share the operations among cells, on top of which the same cell is used in the searched GNN, which is dubbed LRGNN-Repeat in this paper.

\subsubsection{Architecture search from the search space.}
The architecture design can be formulated as the bi-level optimization problem as shown in the following:
\begin{align}
	\min\nolimits_{\bal \in \mathcal{A}} & \;
	\mathcal{L}_{\text{val}} (\bW^*(\bal), \bal), 
	\nonumber
	\\\
	\text{\;s.t.\;} \bW^*(\bal) 
	& = \argmin\nolimits_\bW \mathcal{L}_{\text{train}}(\bW, \bal),
	\nonumber
\end{align}
$ \mathcal{A}$ represents the search space, and $\bal$ represents the neural architecture in $ \mathcal{A}$. $\bW$ represents the parameters of a model from the search space, and $ \bW^*(\bal) $ represents the corresponding operation parameter after training. $\mathcal{L}_{\text{train}}$ and $\mathcal{L}_{\text{val}}$ represent the training and validation loss, respectively.
In this paper, $\bal=\{\bal_c, \bal_m, \bal_r\}$ has three components need to be designed.
As shown in Figure~\ref{fig-cell-framework} (c), $\bal^{(c,i,j)}_c \in \mathbb{R}^{|\mathcal{O}_c|}$ represent the learnable connections between the operation $i$ and $j$ in $c$-th cell. 
$\bal^{(c,i)}_m\in \mathbb{R}^{|\mathcal{O}_m|}$ represent the merge operation used in the aggregation operation $i$ in $c$-th cell.
The LRGNN-Full variant adopts $c=1$ merely, while LRGNN-Repeat share the parameters $\bal^{(i,j)}_c$ and $\bal^{(i)}_m$ cross cells.

The differentiable method is employed in this paper which is widely used in the NAS methods~\cite{liu2018darts,xie2018snas}. We adopted the Gumbel Softmax to relax the search space directly. As shown in the following: 
\begin{align}
	\label{eq-gumble-softmax}
	c_k &=g(\mathcal{O},\bal, k)=\frac{\exp((\log\bm{\alpha}_k + \textbf{G}_k)/\lambda)}{\sum_{j=1}^{\left| \mathcal{O}\right|} \exp((\log\bm{\alpha}_j + \textbf{G}_j)/\lambda)}, \\ \nonumber
\end{align}
$\textbf{G}_k=-\log(-\log(\textbf{U}_k ))$ is the Gumble random variable, and $\textbf{U}_k$ is a uniform random variable, $\lambda$ is the temperature of softmax. 
As shown in Figure~\ref{fig-cell-framework} (c), we use the aggregation operation $3$ as an example, the output can be represented as 
\begin{align}
	\label{eq-connection-results}
	\bH^{(i,3)} &= \sum_{k=1}^{|\mathcal{O}_c|} c_ko_k(\bH^{i}), c_k = g(\mathcal{O}_c, \bal^{(1,i,3)}_c, k), \\
	\label{eq-merge-results}
	\bH^3_{in} &= \sum_{k=1}^{|\mathcal{O}_m|} c_ko_k(\{\bH^{(i,3)}| 0\leq i \leq 2\}), c_k = g(\mathcal{O}_m, \bal^{(1,3)}_m,k),\\
	\label{eq-agg-results}
	\bH^3 &= \text{AGG}(\bA, \bH^3_{in}).
\end{align}
$\bH^i$ is the output feature matrix of operation $i$, and $\bH^{(i,3)}$ represent the results collected based on the connection from operation $i$ to operation $3$. Then, these selected features are merged and the results are denoted as $\bH^3_{in}$, on top of which one aggregation operation \text{AGG} can be applied and the output of operation $3$, i.e., $\bH^3$, will be generated.
The graph representation can be generated as shown in the Alg.~\ref{algo-graph-representation} in Appendix~\ref{sec-appendix-algos}.
We optimize the parameters $\bal$ and $\bW$ with the gradient descent, and the details are provided in Alg.~\ref{algo-optimization} in Appendix~\ref{sec-appendix-algos}. After the training finished, we obtained the searched GNNs by preserving the operations with the largest weights.

In summary, LRGNN aims to design the inter-layer connections in deep stacking-based GNNs to capture the long-range dependencies in the graph classification task. Compared with existing methods which use shallow GNNs or only design deep GNNs in the node classification task, LRGNN evokes the importance of deep stacking-based GNNs in the graph classification task, and the rationality has been justified in Section~\ref{sec-method-twoneeds}. 
Besides, the methods which aim to design adaptive aggregation operations are orthogonal with LRGNN, and they can be incorporated into this paper directly. 
In the following, we empirically evaluate the rationality and effectiveness of LRGNN.

\section{Experiments}
We evaluate the proposed LRGNN against a number of SOTA methods and widely used GNNs, with the goal of answering the following research questions:

\noindent\textbf{Q1:} How does the over-smoothing problem and two design needs empirically affect the performance of stacking-based GNNs in the graph classification task? (Section~\ref{sec-oversmooth})?

\noindent\textbf{Q2:} How does the proposed LRGNN compare to other methods when extracting the long-range dependencies (Section~\ref{sec-performance-comparisons})?

\noindent\textbf{Q3:} How does each component affect the method performance, i.e., the merge and readout operations used in the search space, and the cell-based search space designed considering the search efficiency (Section~\ref{sec-ablation}).


%
%

\subsection{Experimental Settings}

\subsubsection{Datasets}
As shown in Tab.~\ref{tb-graph-dataset}, 
we use five datasets with different graph diameters from different domains.
NCI1 and NCI109 are datasets of chemical compounds~\cite{wale2006comparison}. 
DD and PROTEINS datasets are both protein graphs~\cite{dobsondistinguishing}. 
IMDB-BINARY dataset is movie-collaboration datasets~\cite{yanardag2015deep}. 

\begin{table}[ht]
	\setlength\tabcolsep{1pt}
	\centering
	\footnotesize
	\caption{Statistics of the datasets from three domains.}
	\vspace{-10pt}
	\label{tb-graph-dataset}
	\begin{tabular}{c|C{28pt}|C{28pt}|C{28pt}|C{32pt}|C{30pt}|c}
		\hline
		Dataset     & \# of Graphs & \# of Feature & \# of Classes & Avg.\# of Nodes & Avg.\#  of Edges & Domain    \\ \hline
		NCI1        & 4,110        & 89            & 2             & 29.87           & 32.3             & Chemistry \\ \hline
		NCI109      & 4,127        & 38            & 2             & 29.69           & 32.13            & Chemistry \\ \hline
		DD          & 1,178        & 89            & 2             & 384.3           & 715.7            & Bioinfo   \\ \hline
		PROTEINS         & 1,113        & 3             & 2             & 39.1            & 72.8             & Bioinfo   \\ \hline
		IMDB-BINARY      & 1,000        & 0             & 2             & 19.8            & 96.5             & Social    \\ \hline
	\end{tabular}
	\vspace{-20pt}
\end{table}

\subsubsection{Baselines}
We provide three kinds of baselines in this paper:

\noindent$\bullet$ For stacking-based GNNs, we adopt three baselines in this paper which use different skip-connection schemes: GCN~\cite{kipf2016semi}, ResGCN~\cite{li2019deepgcns} and GCNJK~\cite{xu2018representation}. We vary the GNN layers in $\{4,8,12,16\}$, and then report the best methods in Tab.~\ref{tb-performance} (More results can be found in Appendix~\ref{sec-appendix-setting}). 

\noindent $\bullet$ We provide three pooling methods used in this task: DGCNN~\cite{zhang2018end} baseline with $8$-layer stacked GCN operations and the designed readout function;  $2$-layer SAGPool~\cite{lee2019self} and DiffPool~\cite{ying2018hierarchical} baselines in which each layer has one GCN operation and one pooling operation.

\noindent $\bullet$ Considering the methods which incorporate the higher-order neighbors, we provide four baselines. Firstly, we use the 4-layer stacked GCN, and in each layer nodes aggregate messages from the neighbors exactly one and two hops away. 
This baseline is denoted as TwoHop(L4) in this paper~\cite{abu2019mixhop}. 
For GCNFA(L8)~\cite{alon2020bottleneck} baseline, we use the 8-layer stacked GCN in which the $8$-th layer uses the fully-adjacency matrix instead.  For GCNVIR(L8)~\cite{pham2017graph}, we use the $8$-layer stacked GNN, and in each layer, we add one virtual node. For GraphTrans~\cite{jain2021representing}, we adopt the small variant in this paper which use three GCN layers and four transformer layers.

Compared with these baselines, we provide four LRGNN variants: LRGNN-Full with 8 and 12 aggregation operations, which denoted as B8C1 and B12C1, respectively; LRGNN-Repeat and LRGNN-Diverse with 12 aggregation operations and 3 cells, which denoted as Repeat B12C3 and Diverse B12C3, respectively.


\subsubsection{Implementation details}
For all datasets, we perform 10-fold cross-validation to evaluate the model performance and an inner holdout technique with a 80\%/10\%/10\% training/validation/test for model training and selection. 
In this paper, we set the training and finetuning stages to get the performance for the proposed LRGNN. 
In the training stage, we derived the candidate GNNs from the corresponding search space. 
For the searched GNNs and the baselines, we fine-tune the hyper-parameters of these methods in the finetuning stage.
The details are provided in Appendix~\ref{sec-appendix-setting}.



\subsection{The Rationality of LRGNN}
\label{sec-oversmooth}

\begin{table*}[ht]
\small
	\centering
	\caption{Performance comparisons. We first show the average diameters of the datasets.
		Then, we report the mean test accuracy and the standard deviations based on the 10-fold cross-validation data. For the proposed method, BICJ represent these architectures contains $i$ aggregation operations and $j$ cells. $[Li]$ represents the model depth is $i$.  ``OOM'' means out of memory. The best result in each group is underlined, and the best result in this dataset is highlighted in gray. The average rank on all datasets is provided and the Top-3 methods are highlighted. For GCNFA and GraphTrans, the averaged rank is calculated based the other four datasets.
	}
\vspace{-10pt}
	\label{tb-performance}
\begin{tabular}{c|c|c|c|c|c|c|c}
	\hline
	& Method        & NCI1                                           & NCI109                                         & DD                                             & PROTEINS                                       & IMDB-BINARY                                    & Avg. rank                   \\ \hline
	&Diameter               & 13.33                                          & 13.13                                          & 19.90                                          & 11.57                                          & 1.86                                           &                             \\ \hline
	& GCN           & 78.98(3.24) [L16]                               &  76.64(1.04) [L16]                       & 75.63(2.95) [L16]                              & 75.11(4.51) [L8]                               & 73.80(5.13) [L4]                               & 7.4                         \\ 
	& ResGCN        &  77.88(2.24) [L12]                        & {\ul 76.71(1.83) [L16]}                              & {\ul 76.65(2.73) [L8]}                         & 75.11(3.22) [L8]                               & 73.70(5.70) [L12]                              & 7.2                         \\ 
	\multirow{-3}{*}{Stacking}     & GCNJK         & {\ul 79.24(2.11) [L12] }                             & 75.91(3.61) [L12]                              & 73.16(5.12) [L8]                               & {\ul 75.24(4.15) [L8]}                         & {\ul 74.20(3.76) [L16]}                        & 6.6                         \\ \hline
	& DGCNN(L8)     & {\ul 76.08(1.03)}                              & {\ul 74.58(3.99)}                              & 61.63(5.33)                                    & 73.95(3.04)                                    & 71.70(3.65)                                    & 11.6                        \\ 
& SAGPool       & 72.23(3.68)                                    & 69.78(2.26)                                    & 70.52(5.48)                                    & 71.89(4.03)                                    & {\ul 73.40(3.02)}                              & 12                          \\ 
\multirow{-3}{*}{Pooling}      
& DiffPool      & 75.04(1.98)                                    & 71.48(2.46)                                    & {\ul 77.85(3.53)}                              & {\ul 75.11(2.14)}                              & 72.27(4.47)                                    & 9.2                         \\ \hline
	& TwoHop(L4)    & 76.42(3.83)                                    & 75.33(2.15)                                    & {\ul 74.53(5.24)}                              & {\ul 75.30(4.27)}                              & 72.40(5.61)                                    & 8                           \\ 
	& GCNVIR(L8)    & 70.32(4.32)                                    & 67.99(2.97)                                    & 71.98(4.34)                                    & 71.08(3.14)                                    & 71.50(4.90)                                    & 13.2                        \\ 
	& GCNFA(L8)     & 76.39(3.72)                                    & 73.93(2.87)                                    & OOM                                            & 74.31(4.16)                                    & 73.60(5.62)                                    & 10.25                       \\ 
	& GraphTrans    & {\ul 82.04(1.43)}                              & {\ul 80.18(1.97)}                              & OOM                                            & 75.12(4.89)                                    & {\ul 74.00(5.74)}                              & 4.75                        \\ \hline
	\multirow{-5}{*}{\begin{tabular}[c]{@{}c@{}}Higher-\\ order\end{tabular}} 
	& Full B8C1     & 81.82(1.74) [L4]                               & \cellcolor[HTML]{C0C0C0}{\ul 81.39(1.92) [L6]} & 78.01(3.69) [L6]                               & \cellcolor[HTML]{C0C0C0}{\ul 75.39(4.40) [L6]} & \cellcolor[HTML]{C0C0C0}{\ul 76.20(5.18) [L4]} & \cellcolor[HTML]{C0C0C0}1.6 \\ 
	& Full B12C1    & \cellcolor[HTML]{C0C0C0}{\ul 82.51(1.37) [L7]} & 80.64(2.46) [L7]                               & \cellcolor[HTML]{C0C0C0}{\ul 78.18(2.02) [L9]} & 75.29(4.51) [L3]                               & 74.50(3.31) [L5]                               & \cellcolor[HTML]{C0C0C0}2.4 \\ 
	& Repeat B12C3  & 81.31(1.68) [L6]                               & 80.79(1.24) [L6]                               & 77.25(2.90) [L6]                               & 75.30(4.77) [L3]                               & 74.20(5.41) [L6]                               & \cellcolor[HTML]{C0C0C0}3.6 \\ 
	\multirow{-4}{*}{LRGNN}        & Diverse B12C3 & 80.97(2.73) [L6]                               & 80.98(1.96) [L9]                               & 77.67(3.35) [L5]                               & 74.93(5.15) [L7]                               & 74.20(3.35) [L5]                               & 4.8                         \\ \hline
\end{tabular}
\end{table*}

\subsubsection{The importance of sufficient model depth}
\begin{figure}[ht]
	\centering
	\includegraphics[width=0.45\linewidth]{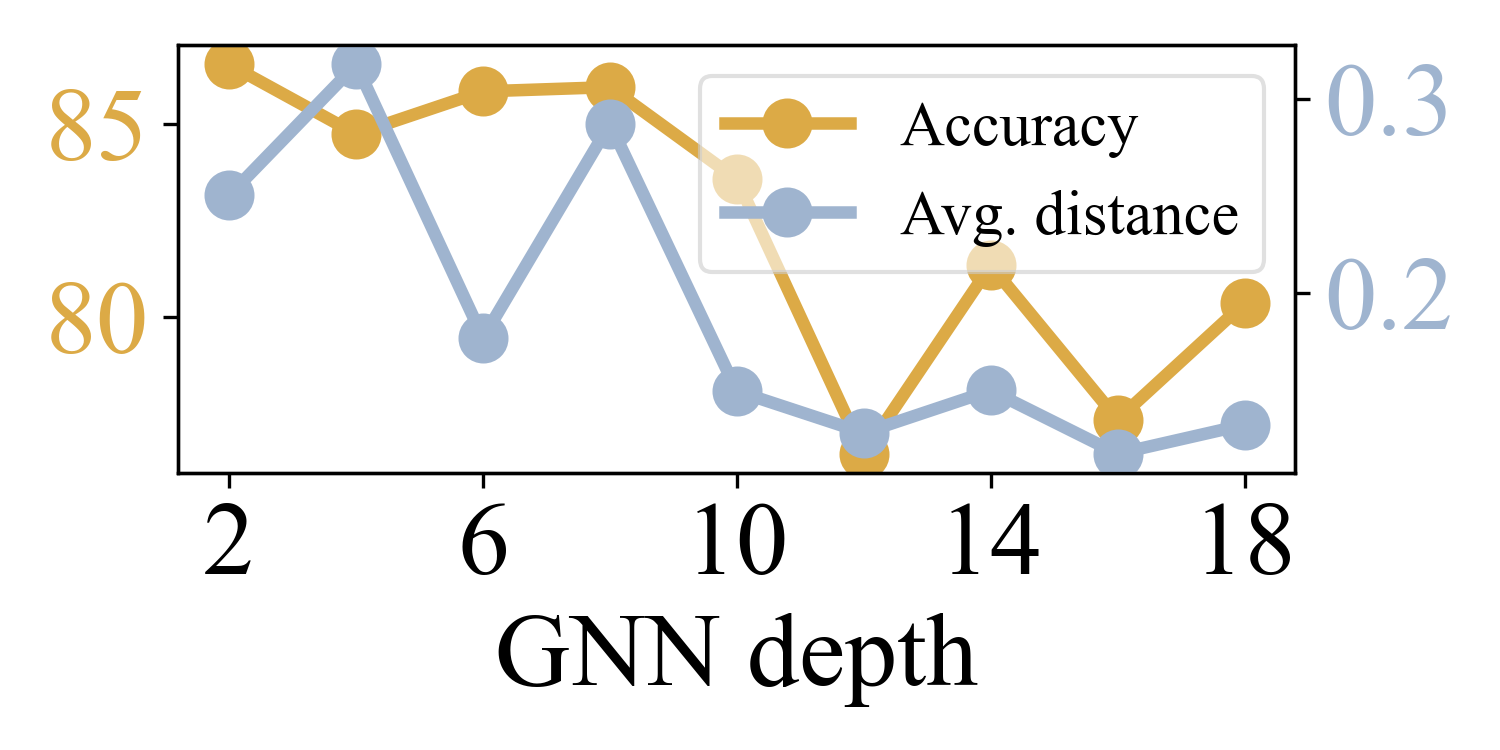}
	\includegraphics[width=0.45\linewidth]{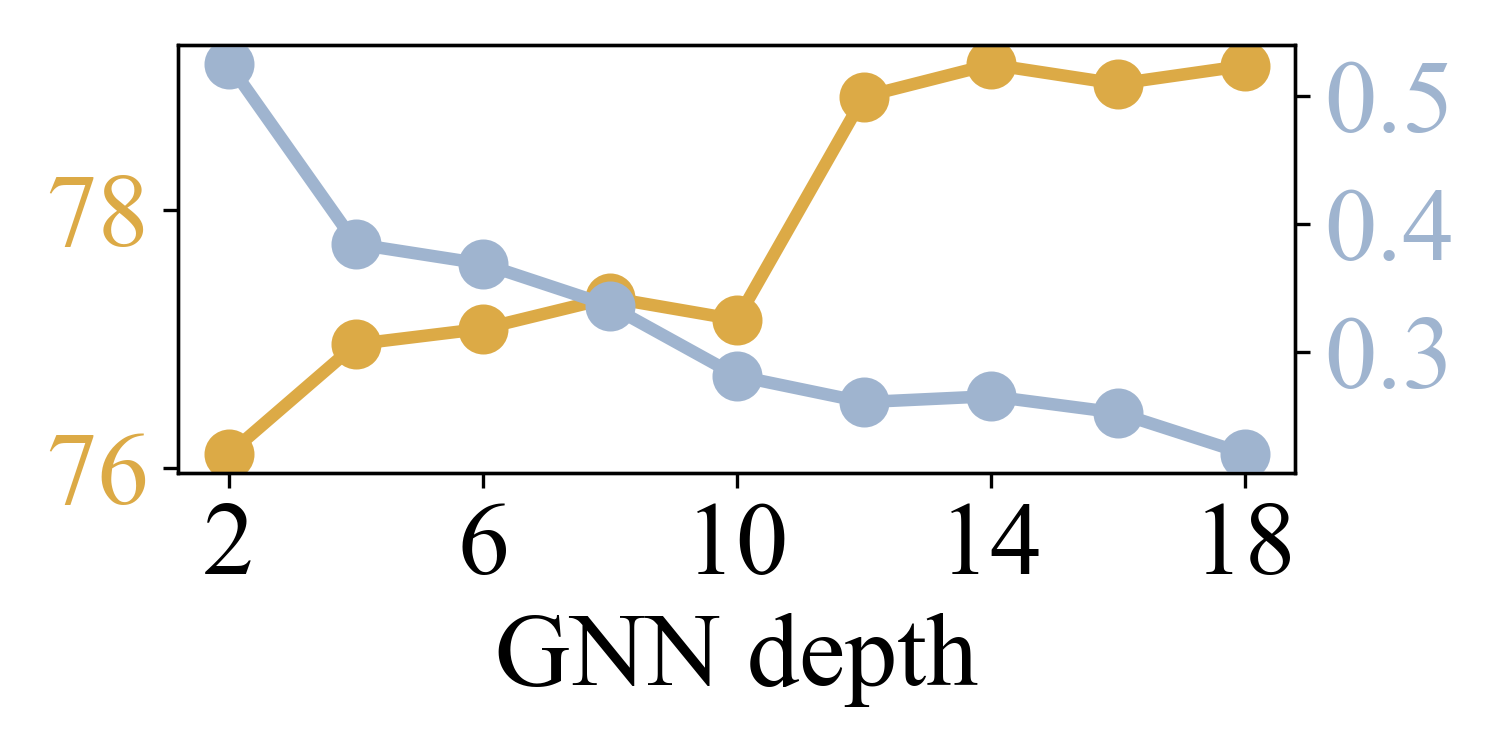}
	\vspace{-10pt}
	\caption{The test accuracy and the averaged distance comparisons on the node classification dataset Cora (Left) and graph classification dataset NCI1 (Right).  }
	\label{fig-smooth-acc}
	\vspace{-15pt}
\end{figure}

In this paper, we use the averaged node pair distance, which is calculated based on $D(h_i, h_j) = \Vert \frac{h_i}{\Vert h_i \Vert_1} -  \frac{h_j}{\Vert h_j \Vert_1} \Vert_1$, to evaluate the over-smoothing problem. 
As shown in Figure~\ref{fig-smooth-acc}, the lower distance values are observed on two datasets as the network goes deeper, which demonstrates the smoother features between node pairs are observed and the over-smoothing problem appeared in both the node classification and graph classification tasks. 
However, the upward trend of model performance is apparent in the NCI1 dataset, while the opposite trend is observed in the Cora dataset. It indicates the smaller influence of the over-smoothing problem on the graph classification task compared with the node classification task. Therefore, stacking sufficient layers to obtain the long-range dependencies is a feasible way.

\subsubsection{The necessity of adaptive skip-connections}
\begin{figure}[ht]
	\vspace{-10pt}
	\centering
	\includegraphics[width=0.4\linewidth]{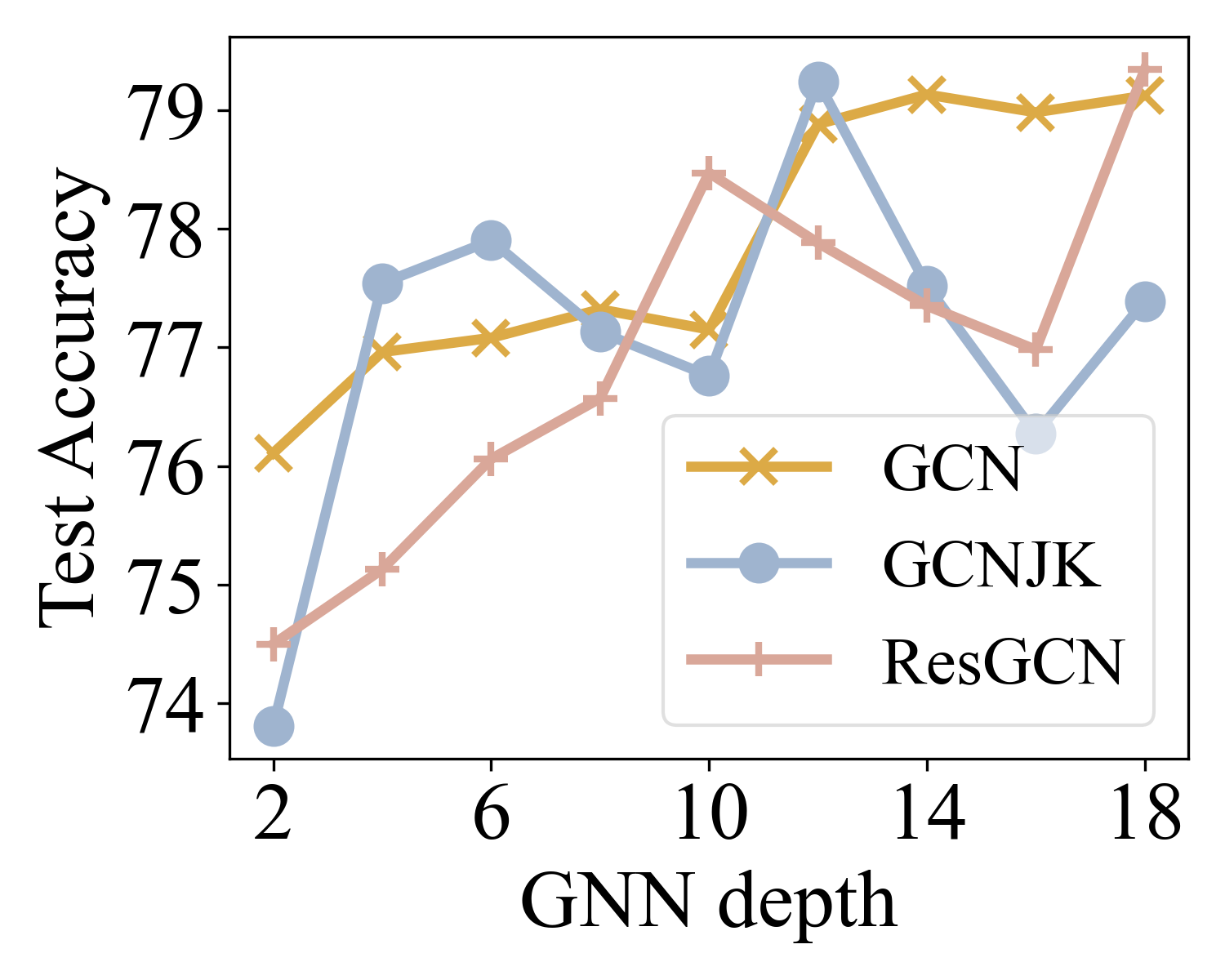}
	\includegraphics[width=0.4\linewidth]{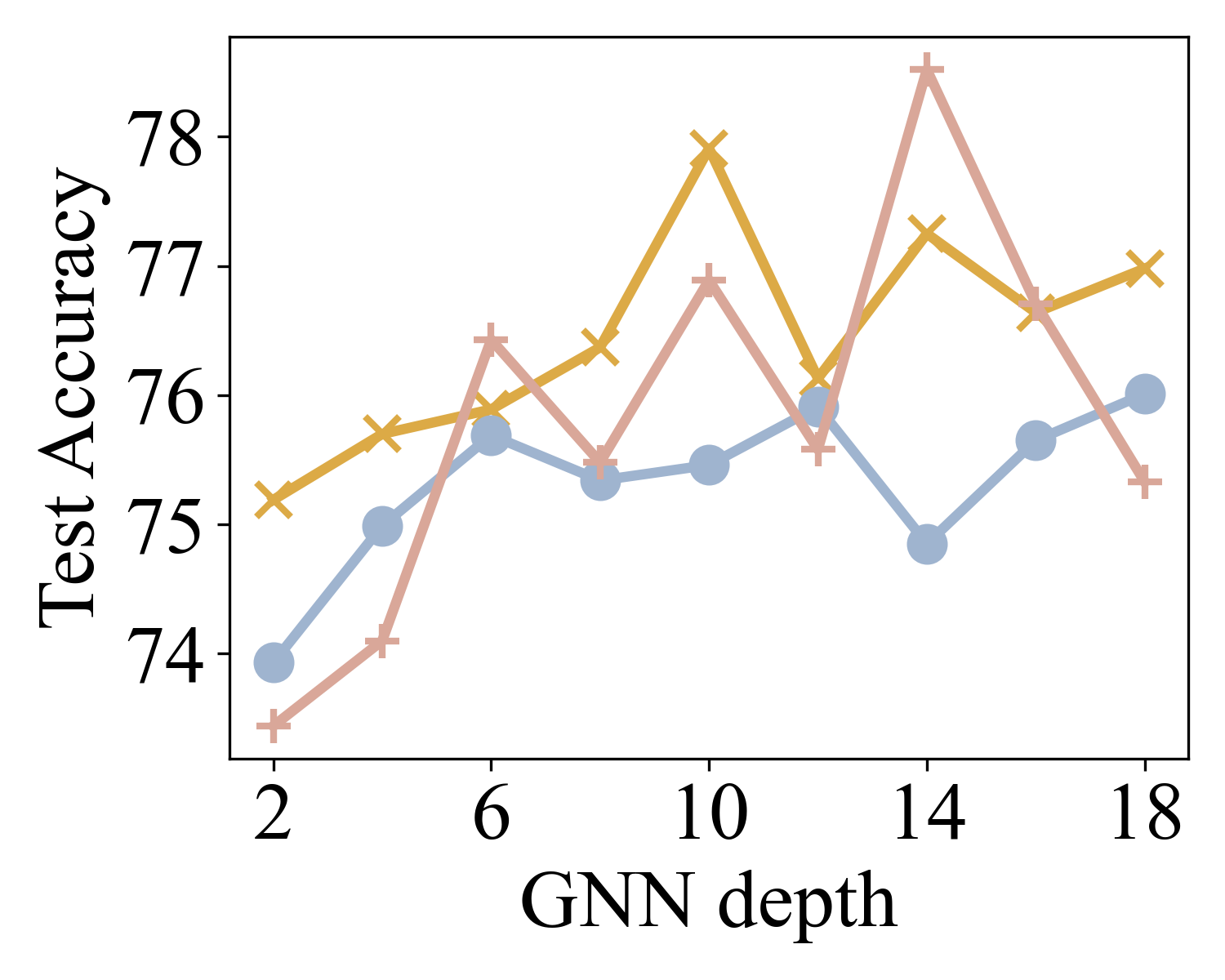}
	\vspace{-10pt}
	\caption{The influence of different skip-connection schemes on NCI1 (Left) and NCI109 (Right) datasets.}
	\label{fig-skipconn-acc}
	\vspace{-15pt}
\end{figure}

As mentioned in Section~\ref{sec-adaptive-connection}, designing adaptive skip-connections can enhance the information utilization from different ranges of neighbors.
In this part, we empirically evaluate the skip-connection schemes on two real-world datasets.
As shown in Figure~\ref{fig-skipconn-acc}, GNNs with different skip-connection schemes have diverse performances in each dataset. Besides, the SOTA performance is achieved by designing the skip-connections, i.e., ResGCN achieves the SOTA performance on the NCI109 dataset and GCNJK ranks first on the NCI1 dataset. 
Given the fixed GNN depth, these three baselines have different ranks on two datasets.
All these results demonstrate the necessity of data-specific skip-connection schemes.

In summary, sufficient GNN depth and adaptive skip-connection schemes are empirically important for stacking-based GNNs in extracting the long-range dependencies. In the following, we compare the proposed LRGNN with the SOTA methods to show its effectiveness.

\subsection{The Effectiveness of LRGNN}
\label{sec-performance-comparisons}
As shown in Table~\ref{tb-performance},  the variants B8C1, B12C1 and LRGNN-Repeat B12C3 can outperform all baselines, which demonstrates the effectiveness of the LRGNN method by designing the inter-layer connections in GNNs to utilize the long-range dependencies adaptively.
Besides, for these methods which achieve the best results on these datasets (the results highlighted in gray in Tab.~\ref{tb-performance}), we observe that the model depth in these methods are very close to the graph radius (diameter/$2$). 
This observation highlights the significance of sufficient model depth in GNNs.
LRGNN-Repeat and LRGNN-Diverse variants constrain the search space and group the aggregation operations into different cells.
Compared with LRGNN-Full, these two variants have fewer parameters based on the same aggregation operations (see Tab.~\ref{tb-cell-memory}), which enables the explorations on much deeper GNNs. However, they have limited search space, which may filter out the expressive architectures and lead to a performance decrease as shown in Tab.~\ref{tb-performance}.

We visualize the searched inter-layer connections of LRGNN-Full B8C1 in Figure~\ref{fig-searched}, and more results can be found in Appendix~\ref{sec-appendix-searched}. It is obvious that different inter-layer connection schemes are obtained on different datasets, and each operation has its own preference for turning on the connections from previous operations. 
The connections between the consecutive layers are emphasized in the NCI109 dataset, while the connections between the inconsecutive layers are widely used in the NCI1 dataset. We highlight one of the longest paths in the searched GNNs,  and the searched architecture on the NCI109 datasets is deeper than that on the NCI1 dataset.
Combine with the SOTA performance in Tab.~\ref{tb-performance}, the effectiveness of designing inter-layer connections is obvious.

\begin{figure}[ht]
	\centering
	\includegraphics[width=0.55\linewidth]{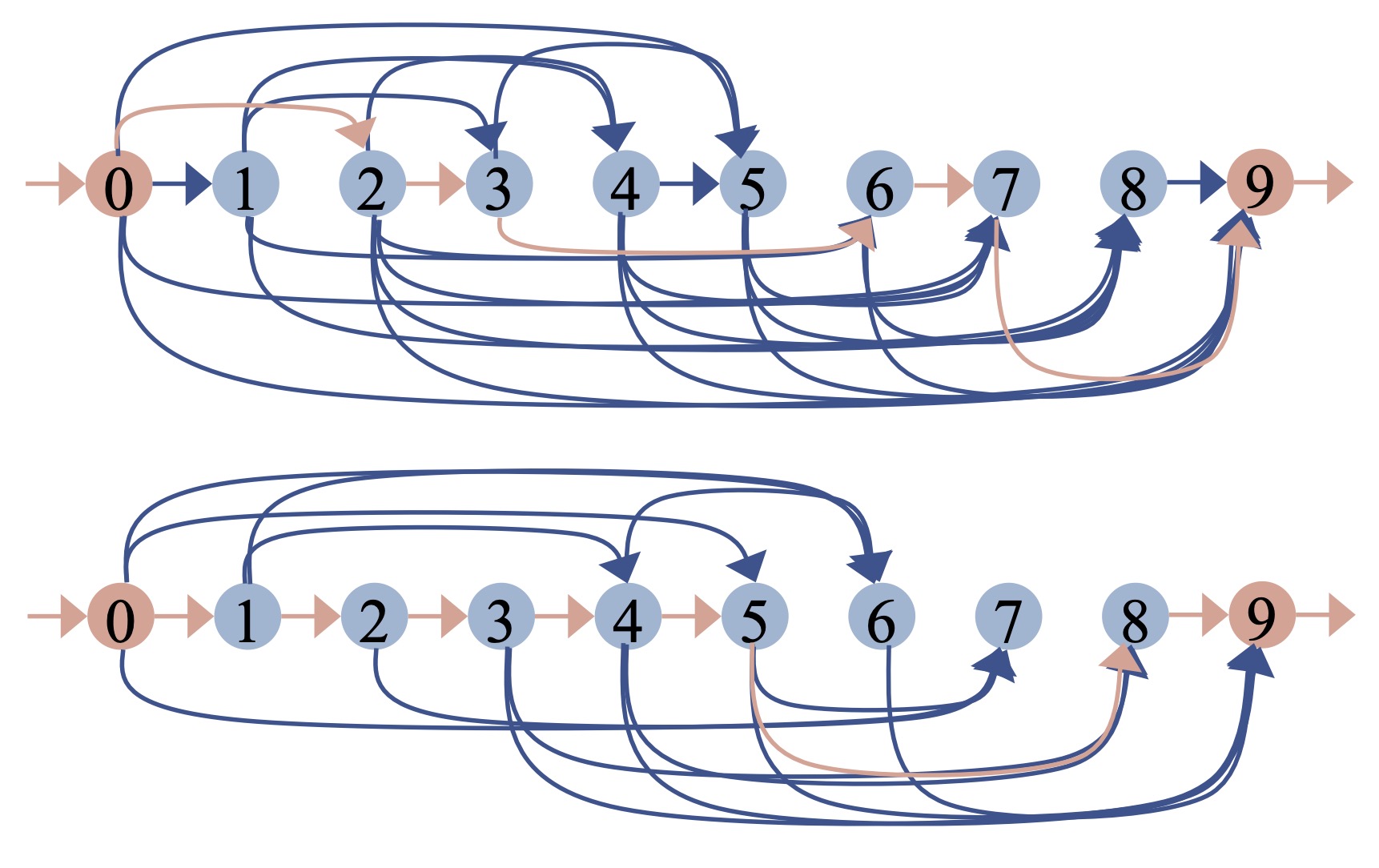}
	\vspace{-10pt}
	\caption{The searched inter-layer connections of LRGNN-Full B8C1 on NCI1 (Top) and NCI109 (Bottom) datasets. We highlight one of the longest path in the GNNs, on top of which the model depth can be obtained.}
	\label{fig-searched}
		\vspace{-10pt}
\end{figure}

Compared with the stacking-based methods,  we observe that three baselines could achieve higher performance with more GNN layers. Besides, by utilizing different skip-connection schemes, ResGCN and GCNJK achieve higher performance than GCN baseline which has no skip-connections. The performance benefits a lot from the inter-layer connections, and the proposed LRGNN achieves better performance by designing the inter-layer connections adaptively.
As to the other two kinds of baselines, GraphTrans has the strongest ability in modeling the long-range information with the help of the Transformer module, which has been proved in the related methods~\cite{jain2021representing,ying2021transformers}. 
Besides, the global pooling method DGCNN and two hierarchical pooling methods, i.e., SAGPool and DiffPool, have different performances on these datasets, and no absolute winner from these two kinds of pooling methods on all datasets, which is consistent with existing methods~\cite{degio2020rethinking,errica2019fair,wei2021pooling}. The grouping-based methods, e.g., DiffPool, 
achieves higher ranks than the selection-based method SAGPool in general. 
Compared with these methods, LRGNN has better performance due to the strong ability of the stacking-based GNNs in preserving the graph structures.

\vspace{-5pt}
\subsection{Ablation Study}
\label{sec-ablation}
To evaluate the contributions of different components, in this section, we conduct ablation studies on the cell-based search space, and two operations added to improve the model expressiveness. 
Considering the limited space, we show the ablation study experiments on aggregation operation and the search algorithm in Appendix~\ref{sec-appendix-ablation}.

\begin{table}[ht]
	\centering
	\footnotesize
	\vspace{-5pt}
	\caption{Evaluations on the cell-based architecture. 
		We show the comparisons of the performance, the parameter numbers(M) in the supernet and the search cost (GPU second) in the training stage. }
	\setlength\tabcolsep{3pt}
	\vspace{-10pt}
	\begin{tabular}{c|c|c}
		\hline
		& DD                                                 & PROTEINS                                            \\ \hline
		Full B4C1     & 77.16(2.98) / 0.23M / 469s                          & 75.30(4.67) / 1.01M / 247s                        \\ \hline
		Full B8C1     & {\ul 78.01(3.69) / 0.92M / 1306s}                   & {\ul 75.39(4.40) / 3.62M / 394s}                   \\ \hline
		Full B12C1    & \cellcolor[HTML]{C0C0C0}78.18(2.02) / 2.28M / 1458s & 75.29(4.51) / 9.06M / 573s                         \\ 
		Repeat B12C3  & 77.25(2.90) / 0.24M / 1226s                         & 75.30(4.77) / 1.05M / 500s                         \\ 
		Diverse B12C3 & 77.67(3.35) / 0.68M / 1286s                         & 74.93(5.15) / 2.79M / 545s                         \\ \hline
		Full B16C1    & OOM                                                  & 74.48(3.89) / 18.39M / 879s                        \\ 
		Repeat B16C4  & 77.24(3.59) / 0.25M / 1697s                         & 74.48(4.75) / 1.07M / 622s                         \\ 
		Diverse B16C4 & 77.59(3.24) / 0.94M /1404s                          & \cellcolor[HTML]{C0C0C0}75.56(4.12) / 3.68M / 665s \\ \hline
	\end{tabular}
	\label{tb-cell-memory}
	\vspace{-10pt}
\end{table}

\subsubsection{The evaluation on cell-based search space}
Considering the search efficiency, we provide one cell-based framework which can constrain the search space by designing the connections in cells. To evaluate the efficiency and effectiveness of the cell-based architectures, we varify the cell numbers and the results are shown in Tab.~\ref{tb-cell-memory}
It is obvious that LRGNN-Full cannot work on DD dataset given $16$ aggregation operations in one cell, while it can work when we assign these 16 aggregation operations into $4$ cells. 
With the same GPU, the cell-based LRGNN-Repeat and LRGNN-Diverse raise the ceiling of the GNN depth compared with the LRGNN-Full. Besides, the Repeat and Diverse variants have fewer parameters and search costs, which demonstrates the search efficiency when using cells. As to the performance, these variants have a smaller search space, and they may filter out the expressive architectures. Therefore, they may be outperformed by LRGNN-Full in some cases. As a conclusion, LRGNN-Full can achieve higher-performance while the other two variants have advantages in the training efficiency.

\subsubsection{The evaluation of the merge and readout operations.}
In the search space, we provide the merge operation and readout operation to improve the model expressiveness as mentioned in  Section~\ref{sec-design-space}. Then, we conduct ablation studies on these two operations to show their effectiveness in improving the model performance.

\begin{figure}[ht]
	\centering
		\vspace{-10pt}
	\includegraphics[width=0.9\linewidth]{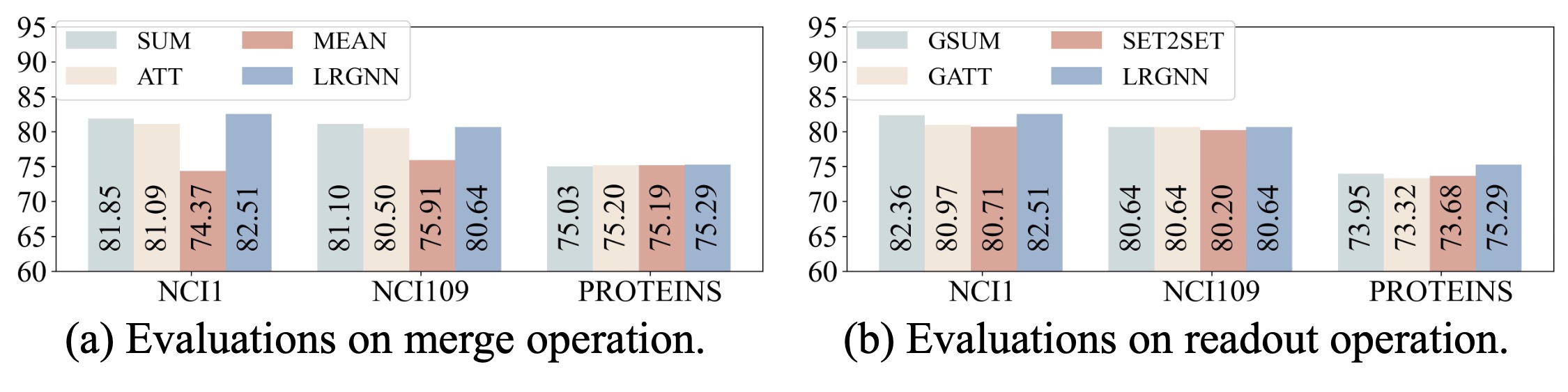}
		\vspace{-10pt}
	\caption{Evaluations on the merge and readout operation. These methods are constructed based on LRGNN-Full B12C1 variant and different merge or readout operations.}
	\label{fig-ablation-readoutmerge}
		\vspace{-10pt}
\end{figure}

In this paper, we provide a set of merge operations that are expected to improve the model expressiveness. 
As shown in Figure~\ref{fig-ablation-readoutmerge} (a), compared with LRGNN which designs the merge operations adaptively, using the pre-defined merge operations may not achieve higher performance in general. The searched results are visualized in Appendix~\ref{sec-appendix-searched}, from which we can observe the different merge operations are utilized. Combine with better performance, the effectiveness to design the merge operations adaptively is obvious.

The readout operation is responsible for generating the graph representation vector for the given graph. 
In this paper, we provide seven candidates and learn to design this operation adaptively. 
As shown in Figure~\ref{fig-ablation-readoutmerge} (b),
LRGNN achieves the SOTA performance in three datasets, and all of them utilize the \texttt{GSUM} readout operation in the searched architectures. 
Besides, the baseline which uses the predefined readout operation \texttt{GSUM} achieves comparable performance as well. 
Nevertheless, using the other two readout operations lead to a performance drop, and then it is important to design the readout operation adaptively.

\vspace{-5pt}
\section{Conclusion}
In this paper,  we provide a novel method LRGNN to capture the long-range dependencies with stacking GNNs in the graph classification task.
%
We justify that the over-smoothing problem has smaller influence on the graph classification task, and then employ the stacking-based GNNs to extract the long-range dependencies. Two design needs, i.e., sufficient model depth and adaptive skip-connections, are provided when designing the stacking-based GNNs.  
To meet these two design needs, we unify them into inter-layer connections, and then design these connections with the help of NAS.
Extensive experiments demonstrate the rationality and effectiveness of the proposed LRGNN.
For future work, we will evaluate the importance of different hops' neighbors in different graph pairs, and then extract this information for each graph independently.



\section*{Acknowledgment}
Q. Yao was in part sponsored by NSFC (No. 92270106) and CCF-Tencent Open Research Fund.

\clearpage
\balance
 \bibliographystyle{ACM-Reference-Format}
\bibliography{ref}

\clearpage
\appendix

\section{Method}
\subsection{The Proof of Proposition~\ref{prop-smooth}}
\label{sec-appendix-prop2}
\begin{proof}
	It has been proved in Theorem 1~\cite{li2018deeper} that $(I-L^{sym})^{\infty}\bW$ will converge to the linear combination of eigenvectors of eigenvalue 1. The satisfied graphs will converged to different points, which leads to differnt node features in these two graphs. Then, based on the Lemma 5 in~\cite{xu2018powerful}, unique graph representation vector will be generated for each graph, and then two graphs be distinguished.
\end{proof}

\subsection{ The Proof of Proposition~\ref{prop-long-range}}
\label{sec-appendix-prop1}
\begin{proof}
It has been proved in Theorem 3~\cite{xu2018powerful} that GIN is as powerful as the 1-WL test. For two graphs which can be distinguished by 1-WL test at 4-th iteration, we can obtain that $\mathcal{A}_k^{GIN}(\mathcal{G}_1) \neq \mathcal{A}_k^{GIN}(\mathcal{G}_2)$ and $\mathcal{A}_l^{GIN}(\mathcal{G}_1) = \mathcal{A}_l^{GIN}(\mathcal{G}_2), l \leq k$. Furthermore, the 1-WL test can distinguish two graphs after $k$-th iterations, and then $\mathcal{A}_l^{GIN}(\mathcal{G}_1) \neq \mathcal{A}_l^{GIN}(\mathcal{G}_2)$ also holds when $l>k$.
\end{proof}

We use the graphs in Figure~\ref{fig-long-range-example}(a) as an example.
Two graphs have the same number of nodes, and all nodes have the same feature, e.g., all-one vector.

\noindent$\bullet$ $0$-layer GNN, which is constructed by MLP and a readout operation, cannot distinguish these two graphs. 
These two graphs have the same feature set, and MLPs with any layers will generate the same feature set for them as well.
Based on these two graphs, any simple readout function designed on the node features will generate the same graph representation vectors. For example, take the mean or summation of all node embeddings as the graph-level representation vector. Set2Set operation~\cite{vinyals2015order} and SortPool~\cite{zhang2018end} have the same results.
Therefore, the $0$-layer GNN cannot distinguish these two graphs.

\noindent$\bullet$ Shallow GNN cannot distinguish these two graphs.
As shown in Figure~\ref{fig-long-range-example} (b), we visualize the computation graph, which is denoted as $S1$-$S4$ in the figure, and nodes in each row use the same computation graph. In $2$-th iteration, 1-WL test will generate the same feature for the nodes in the same row. Therefore, 2-layer GIN, which is constructed by stacking two GIN aggregation operations and one readout function mentioned before, cannot distinguish two graphs as well. The $3$-layer GNN has the same situation.

\noindent$\bullet$ $4$-layer GNN can distinguish these two graphs. 
As shown in Figure~\ref{fig-long-range-example} (c), we visualize the computation graph for node $2$ in $\mathcal{G}_2$. It contains the $S1$ and $S3$ graphs, while the nodes in $\mathcal{G}_1$ cannot achieve this. Therefore, two graphs have different feature sets in $4$-iteration in the 1-WL test. Therefore, $4$-layer GIN can distinguish two graphs.

In summary, capturing long-range dependencies is useful for the graph classification task, which is correlated with the 1-WL test.

\vspace{-10pt}
\subsection{Experimental Details for Figure~\ref{fig-diameter-connection}}
\label{sec-appendix-adaptive-exp}
\noindent\textbf{Datasets.} Three datasets are constructed based on the NCI1 dataset which will be introduced in the following. For the first dataset, we sample 10 graphs with diameter $5$ and 50 graphs with diameter $14$. The other two datasets are constructed in the same way. The graph label is determined by the graph diameter. Therefore, these three datasets only have two labels. The diameter distribution for these datasets is provided in Figure~\ref{fig-diameter-connection}.

\noindent\textbf{Baselines.} We provide six GNNs with different inter-layer connections as shown in Figure~\ref{fig-diameter-connection}. Based on GCN aggregation operation, $A1$-$A3$ have different connection schemes between the consecutive layers and lead to different layer numbers, i.e., 2, 5, and 7, respectively. GNN $A4$-$A6$ use the different connections among inconsecutive layers.

\noindent\textbf{Implementation details.} We evaluate each GNN on three datasets. We fix the hidden dimension to 128, and use a batch size of 64, a fixed learning rate 0.01, and Adam optimizer. We visualize the averaged test accuracy in Figure~\ref{fig-diameter-connection} based on 10-fold cross-validation.

\subsection{Limitations of Existing Methods}
\label{sec-appendix-pooling-illustrations}
\begin{figure}[ht]
	\centering
	\includegraphics[width=0.7\linewidth]{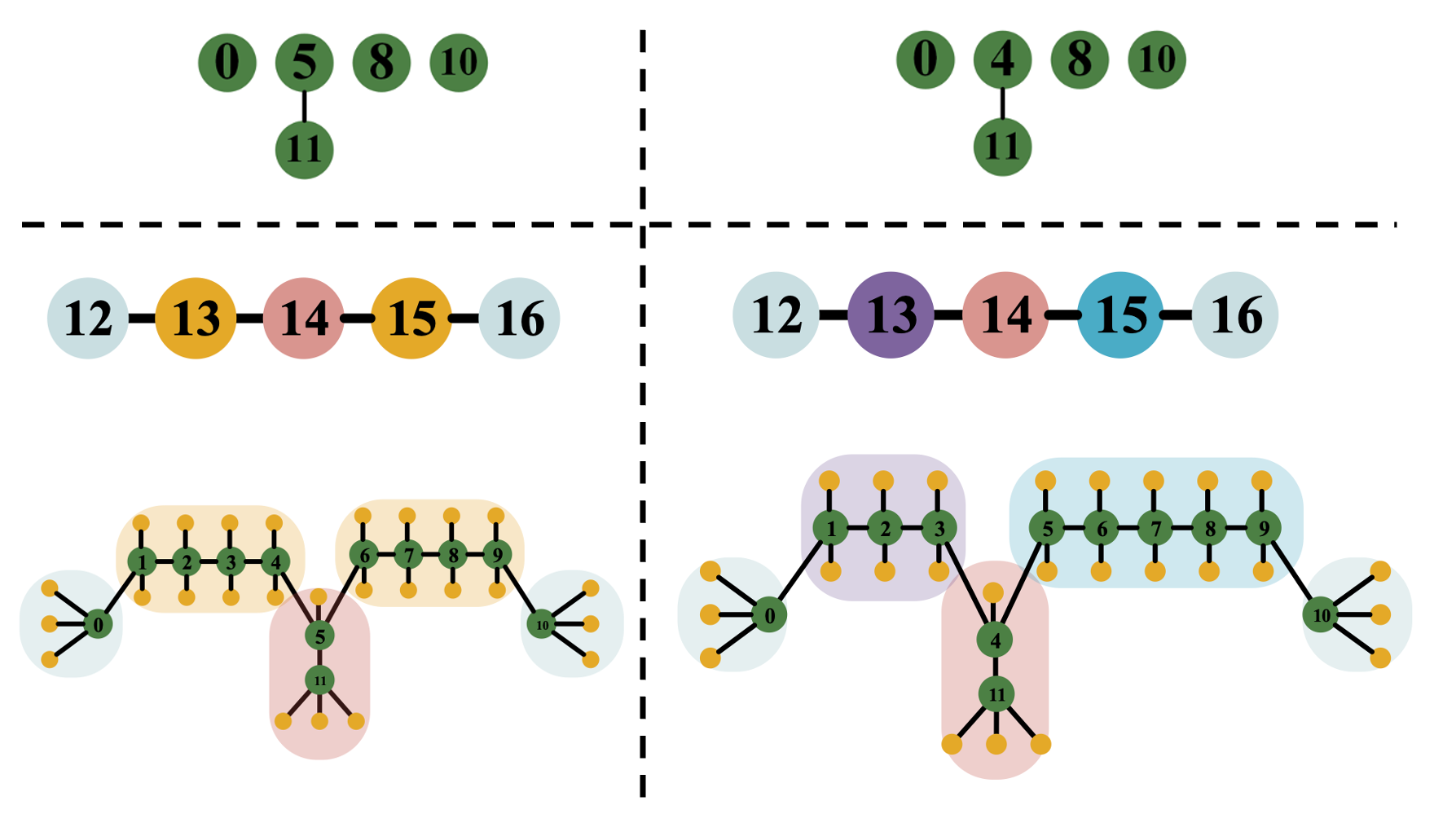}
	\vspace{-10pt}
	\caption{The pooling results of $\mathcal{G}1$ (left) and $\mathcal{G}2$ (right). The coarse graphs in the top row is the results of the selection-based operations, and the bottom row is the results of grouping-based operations. }
	\label{fig-pooling-results}
\end{figure}

Using the graph pairs shown in Figure~\ref{fig-long-range-example} as an example, we visualize the pooling results in Figure~\ref{fig-pooling-results}, in which two different graphs generate the same coarse graph after pooling operations.
The selection-based pooling operations preserve the same node sets in two graphs, and the formulated coarse graphs are the same as the other.
As to the grouping-based operations,  the subgraphs in each colored block have similar features and are densely connected, and they are grouped into the corresponding new nodes in the coarse graph. The same coarse graphs are obtained as well.

\vspace{-10pt}
\subsection{The details of the method}
\label{sec-appendix-algos}
Using the LRGNN-Full as an example, we show the calculation of graph representations in Alg.~\ref{algo-graph-representation}, and show the optimization details in Alg.~\ref{algo-optimization}. 

\begin{algorithm}[ht]
	\caption{Generating the graph representations in LRGNN-Full}
	\label{algo-graph-representation}
\scriptsize
	\begin{algorithmic}[1]
		\Require{Input graph $\mathcal{G}=(\bA, \bH^{input})$, the aggregation number $B$, aggregation operation \text{AGG}.}
		\Ensure{The graph representation $h_{\mathcal{G}}$.}
			\State Update the node feature in the pre-processing stage $\bH^0=f_{pre}(\bH^{input})$.
			\For {$j=1,\ldots, B$}
				\State Calculate the aggregation results $\bH^j$ based on Eq.~\eqref{eq-connection-results},\eqref{eq-merge-results} and \eqref{eq-agg-results}.
			\EndFor
			\State Update the node feature in the post-processing stage $\bH^F=f_{post}(\bH^B)$.
			\State $h_{\mathcal{G}} = \sum_{i=1}^{|O_{r}|}c_i o_i(\bA, \bH^F)$ where $c_k = g(\mathcal{O}_r, \bal_r, k)$ .   $//$ Readout \\
			\Return $h_{\mathcal{G}}$ 
	\end{algorithmic}
\end{algorithm}

\begin{algorithm}[ht]
\scriptsize
	\caption{Designing the inter-layer connections for the stacking-based GNNs.}
	\label{algo-optimization}
	\begin{algorithmic}[2]
		\Require{Training dataset $\mathcal{D}_{train}$, validation dataset $\mathcal{D}_{val}$, the search epoch $T$. }
		\Ensure{The searched GNN.}
		\State Random initialize the architecture parameters $\bm{\alpha}$ and operation parameter $\bW$.
		\While{$t=1,\ldots,T$}
		\For {each minibatch $\mathcal{G}_b \in \mathcal{D}_{train}$}
		\State Calculate the graph representations $h_{\mathcal{G}_b}$ as shown in Alg.~\ref{algo-graph-representation}.
		\State Update $\bW$
		\EndFor
		\For {each minibatch $\mathcal{G}_b \in \mathcal{D}_{val}$}
		\State Calculate the graph representations $h_{\mathcal{G}_b}$ as shown in Alg.~\ref{algo-graph-representation}.
		\State Update $\bm{\alpha}$.
		\EndFor
		\EndWhile
		\State Preserve the candidates with the largest weight in each learnable connection, merge and readout operations.\\
		\Return The searched GNN.
	\end{algorithmic}
\end{algorithm}

\vspace{-10pt}
\section{Experiments}

\subsection{Implementation Details}
\label{sec-appendix-setting}
All models are implemented with Pytorch~\cite{paszke2019pytorch} on a GPU RTX 3090 (Memory: 24GB, Cuda version: 11.2).
We adopt the same data preprocessing manner provided by PyG\footnote{https://github.com/pyg-team/pytorch\_geometric} and split data by means of a stratification technique with the same seed.
We finetune the human-designed architectures and the searched architectures with the help of Hyperopt~\footnote{https://github.com/hyperopt/hyperopt}. In the finetuning stage, each GNN has 20 hyper steps. For LRGNN,  a set of hyperparameters will be sampled from Tab.~\ref{tb-hypers-proposed} in each step, and the baselines will be sampled from Tab.~\ref{tb-hypers-baseline}. Then we select the final hyper-parameters on the validation set, from which we can obtain the final performance of this GNN. 
For stacking-based baselines GCN and ResGCN, we vary the GNN layers in $\{4,8,12,16\}$, and then report the best methods in Tab.~\ref{tb-performance}. The detailed results are shown in Tab.~\ref{tb-performance-stacking}.
Especially, for the GraphTrans baseline, we adopt the provided code~\footnote{https://github.com/ucbrise/graphtrans} and split the data as mentioned before.

\begin{table}[ht]
	\scriptsize
	\centering
	
	\caption{Hyperparameter space in the finetuning stage for the proposed method.}
	\label{tb-hypers-proposed}
	\vspace{-10pt}
	\begin{tabular}{l|l}
		\hline
		Dimension     & Operation                    \\ \hline
		Embedding size   & 8, 16, 32, 64, 128, 256 \\ \hline
		Dropout rate  & 0, 0.1, 0.2, $\cdots$, 0.9          \\ \hline
		Learning rate & $[0.001, 0.025]$             \\ \hline
		Optimizer                          & Adam, AdaGrad                \\ \hline
	\end{tabular}
\end{table}

\vspace{-10pt}
\begin{table}[ht]
	\scriptsize
	\centering
		\caption{Hyperparameter space for human-designed baselines.}
		\vspace{-10pt}
		\label{tb-hypers-baseline}
		\begin{tabular}{l|l}
			\hline
			Dimension             & Operation              \\ \hline
			Global pooling function &  \texttt{GMEAN}, \texttt{GSUM} \\ \hline
			Embedding size      & 8, 16, 32, 64, 128, 256, 512 \\ \hline
			Dropout rate     & 0, 0.1, 0.2,...,0.9    \\  \hline
			Learning rate    & $[0.001, 0.025]$   \\ 
			\hline
		\end{tabular}
\end{table}

\vspace{-10pt}
\begin{table}[ht]
	\scriptsize
	\centering
	\caption{Performance comparisons on the stacking-based baselines GCN, ResGCN and GCNJK with different layers.	}
	\vspace{-10pt}
	\label{tb-performance-stacking}
	\begin{tabular}{c|c|c|c|c|c}
		\hline
		Method      & NCI1         & NCI109       & DD          & PROTEINS    & IMDBB        \\ \hline
		GCN(L4)     & 76.96(3.69)  & 75.70(4.03)  & 73.59(4.17) & 74.84(3.07) & 73.80(5.13)  \\
		GCN(L8)     & 77.32(3.17)  & 76.37(2.62)  & 74.01(2.85) & 75.11(4.51) & 71.60(4.30)  \\
		GCN(L12)    & 78.88(2.76)  & 76.13(2.15)  & 74.53(4.49) & 74.30(3.79) & 73.50(3.17)  \\
		GCN(L16)    & 78.98(3.24)  & 76.64(1.04)  & 75.63(2.95) & 72.42(4.83) & 71.50(4.94)  \\ \hline
		ResGCN(L4)  & 75.13(2.87)  & 74.10(5.183) & 73.85(3.70) & 74.40(5.22) & 73.50(4.83)  \\
		ResGCN(L8)  & 76.57(2.306) & 75.48(3.533) & 76.65(2.73) & 75.11(3.22) & 73.20(6.36)  \\
		ResGCN(L12) & 77.88(2.214) & 75.58(2.027) & 75.38(3.03) & 73.40(5.62) & 73.70(5.70)  \\
		ResGCN(L16) & 76.98(2.907) & 76.71(1.836) & 75.46(3.29) & 73.23(3.89) & 73.19(3.61)  \\ \hline
		GCNJK(L4)   & 77.54(3.47)  & 74.99(3.54)  & 72.40(4.51) & 73.21(2.71) & 73.20(6.14)  \\
		GCNJK(L8)   & 77.13(3.29)  & 75.34(3.14)  & 73.16(5.12) & 75.24(4.15) & 70.91(5.50)  \\
		GCNJK(L12)  & 79.24(2.11)  & 75.91(3.61)  & 71.90(4.87) & 71.25(2.67) & 72.89((5.79) \\
		GCNJK(L16)  & 76.27(3.32)  & 75.65(2.84)  & 69.86(4.40) & 71.16(4.92) & 74.20(3.76)  \\ \hline
	\end{tabular}
\end{table}

\vspace{-10pt}
\subsection{Searched GNNs}
\begin{figure*}[ht]
	\centering
	\includegraphics[width=0.45\linewidth]{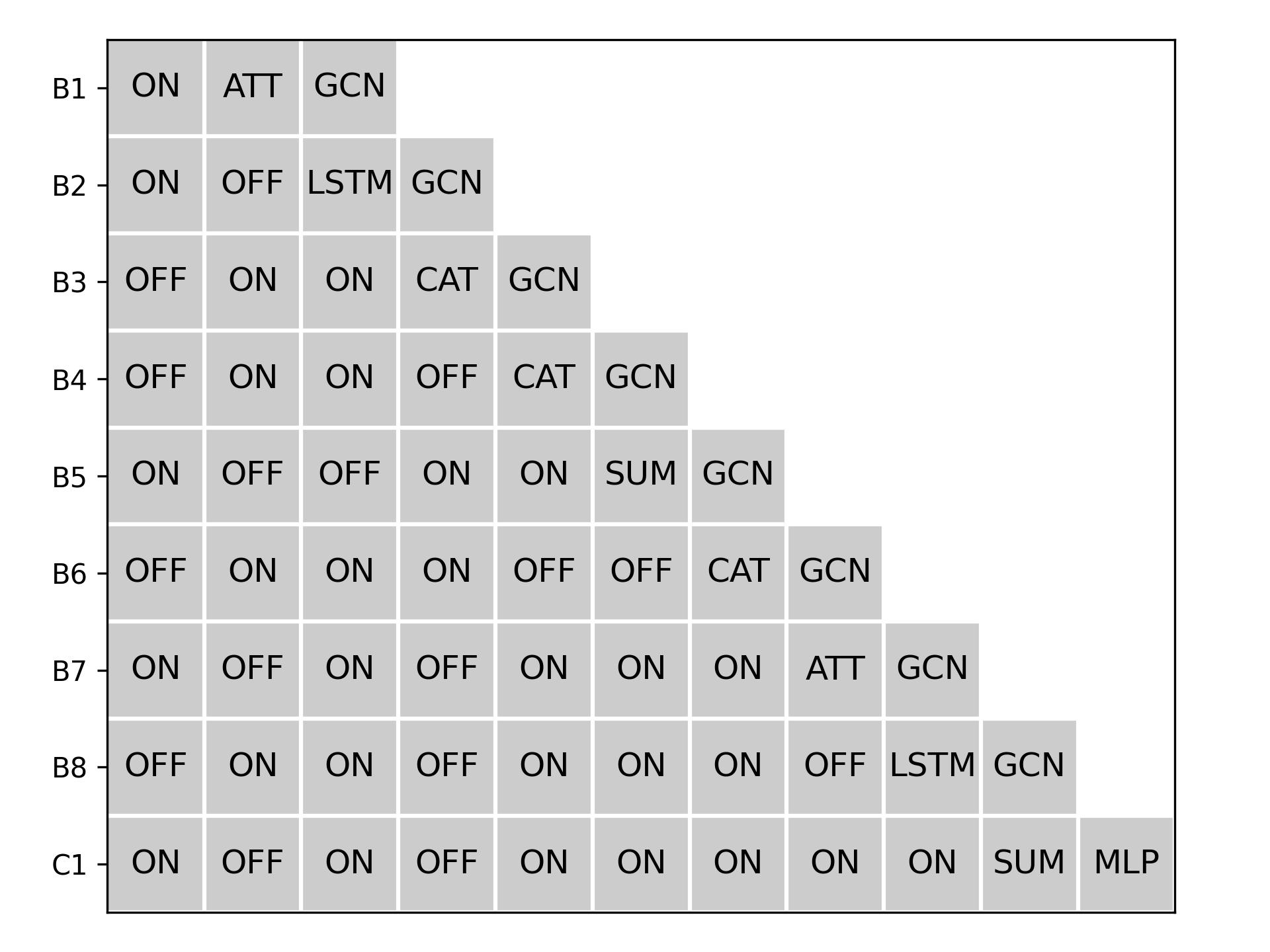}
	\includegraphics[width=0.45\linewidth]{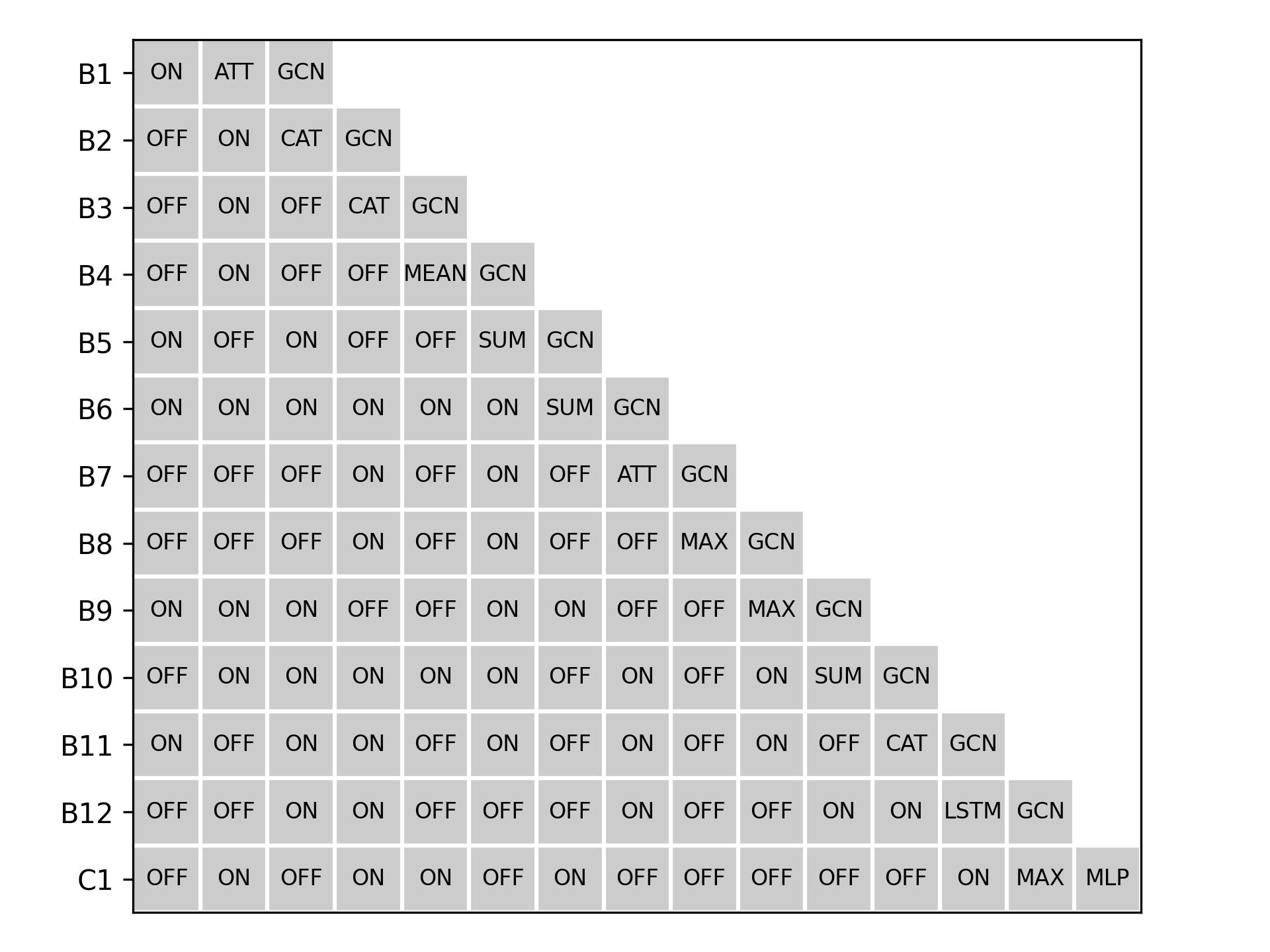}
	\includegraphics[width=0.45\linewidth]{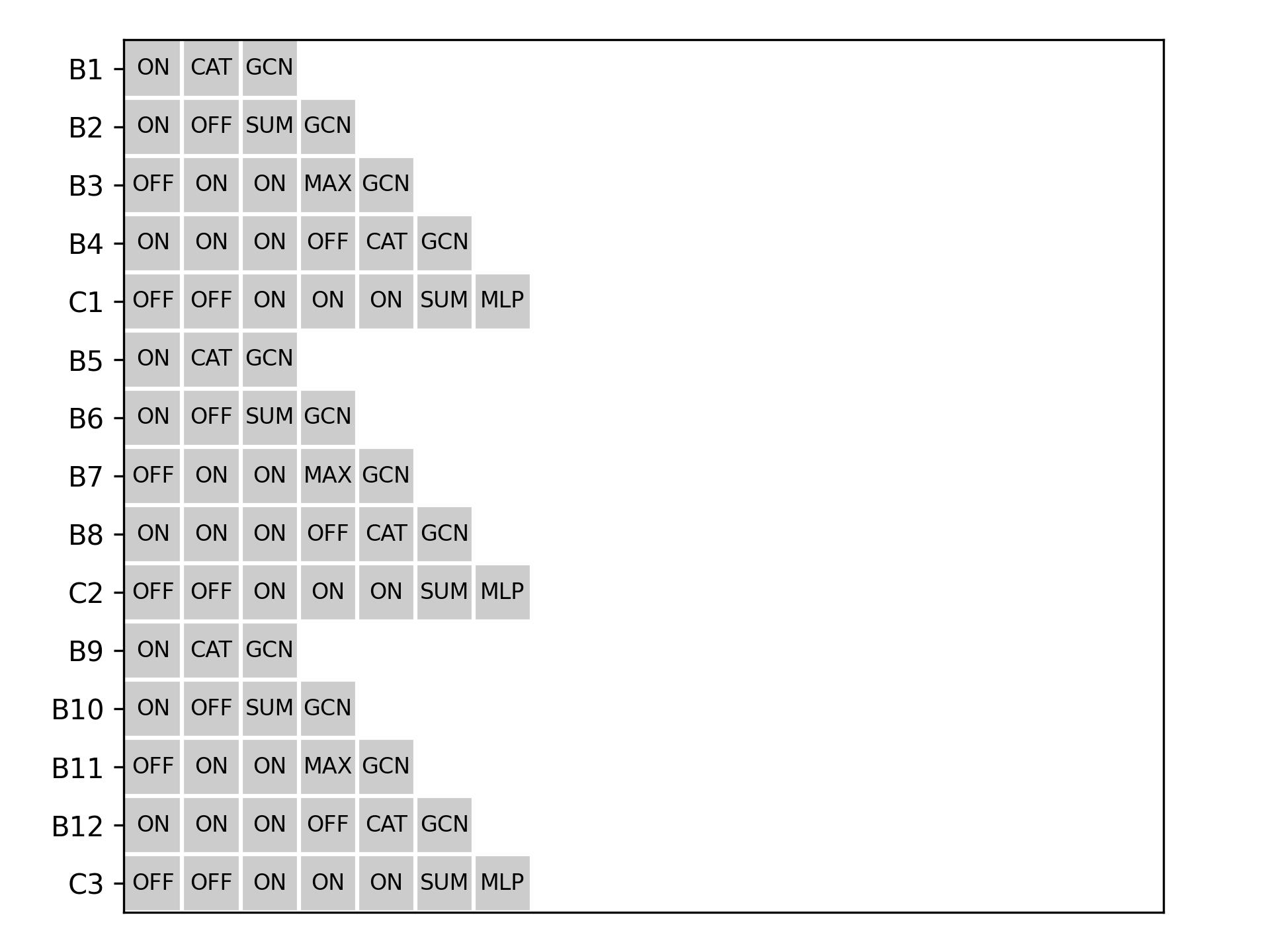}
	\includegraphics[width=0.45\linewidth]{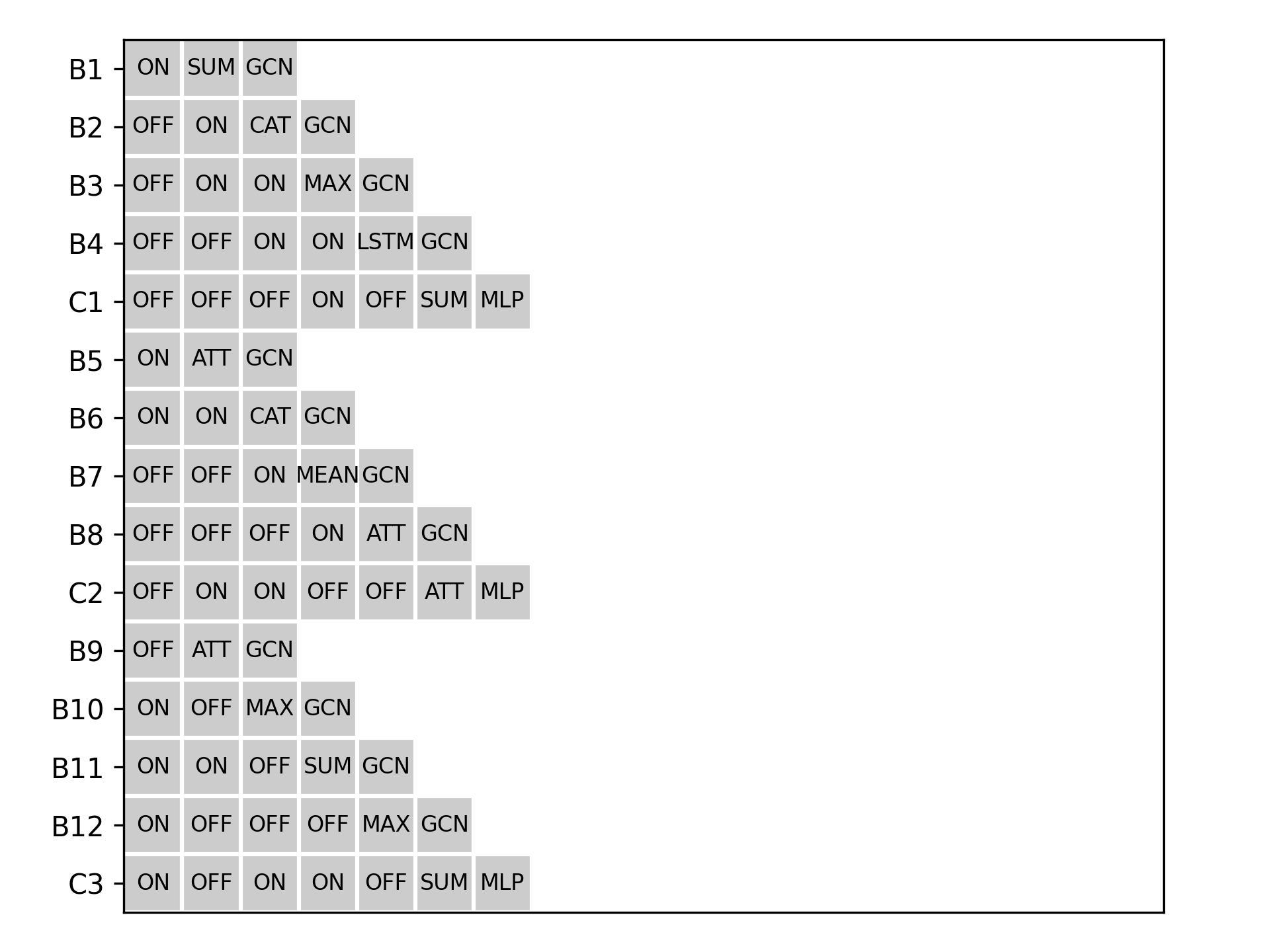}
	\caption{The searched architectures on NCI1 dataset. Four variants from the left to right are the Full-B8C1, Full-B12C1, Repeat-B12C3, and Diverse-B12C3, respectively. In each row, the operations on each connections are showed. Besides, the merge operation and the aggregation operation are shown as well.  In the post-processing operation, which denoted as ``C'' in this figure, use the MLP operation instead.}
	\label{fig-searched-arch-nci1}
\end{figure*}

\begin{figure*}[ht]
	\centering
	\includegraphics[width=0.45\linewidth]{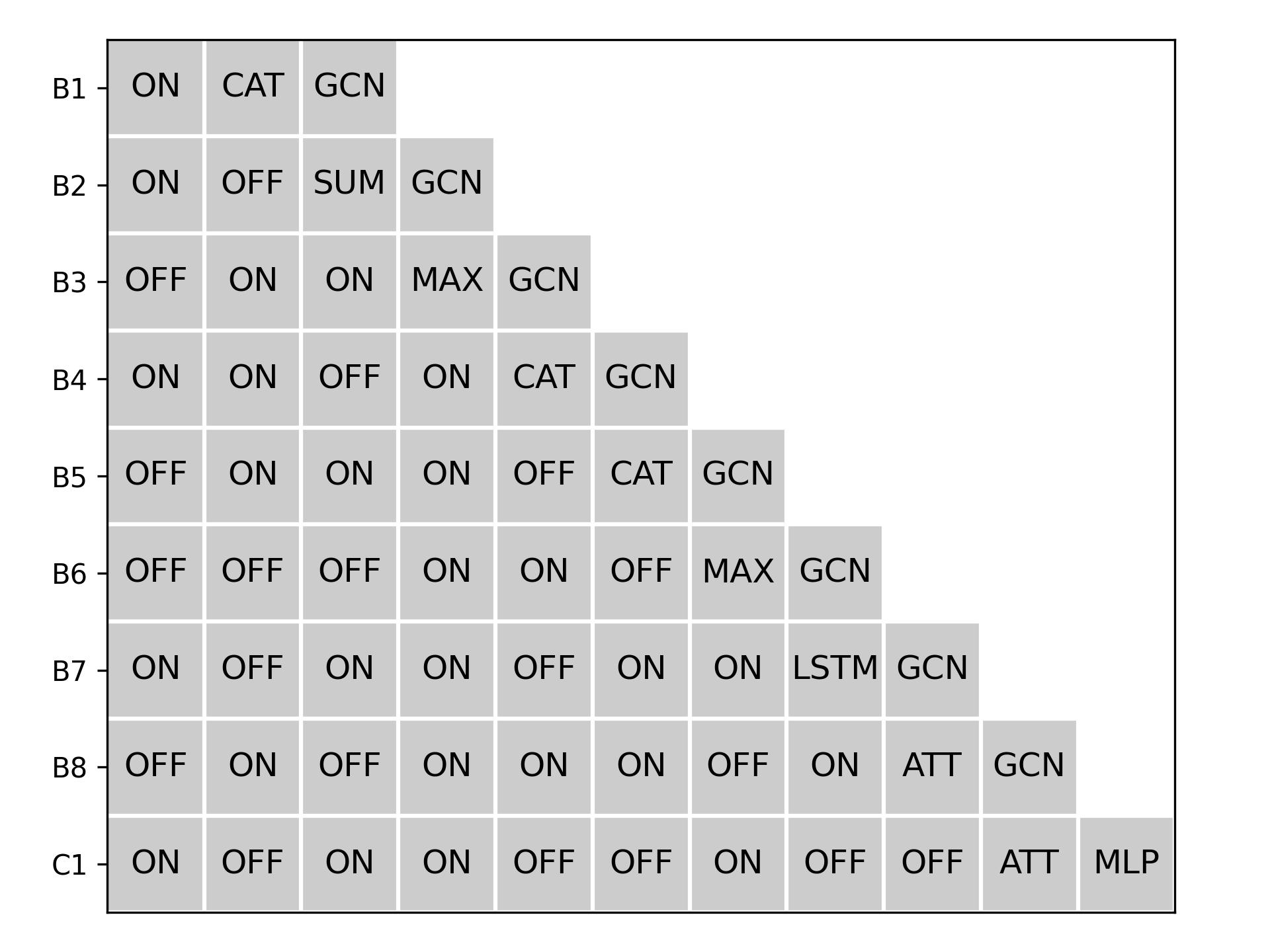}
	\includegraphics[width=0.45\linewidth]{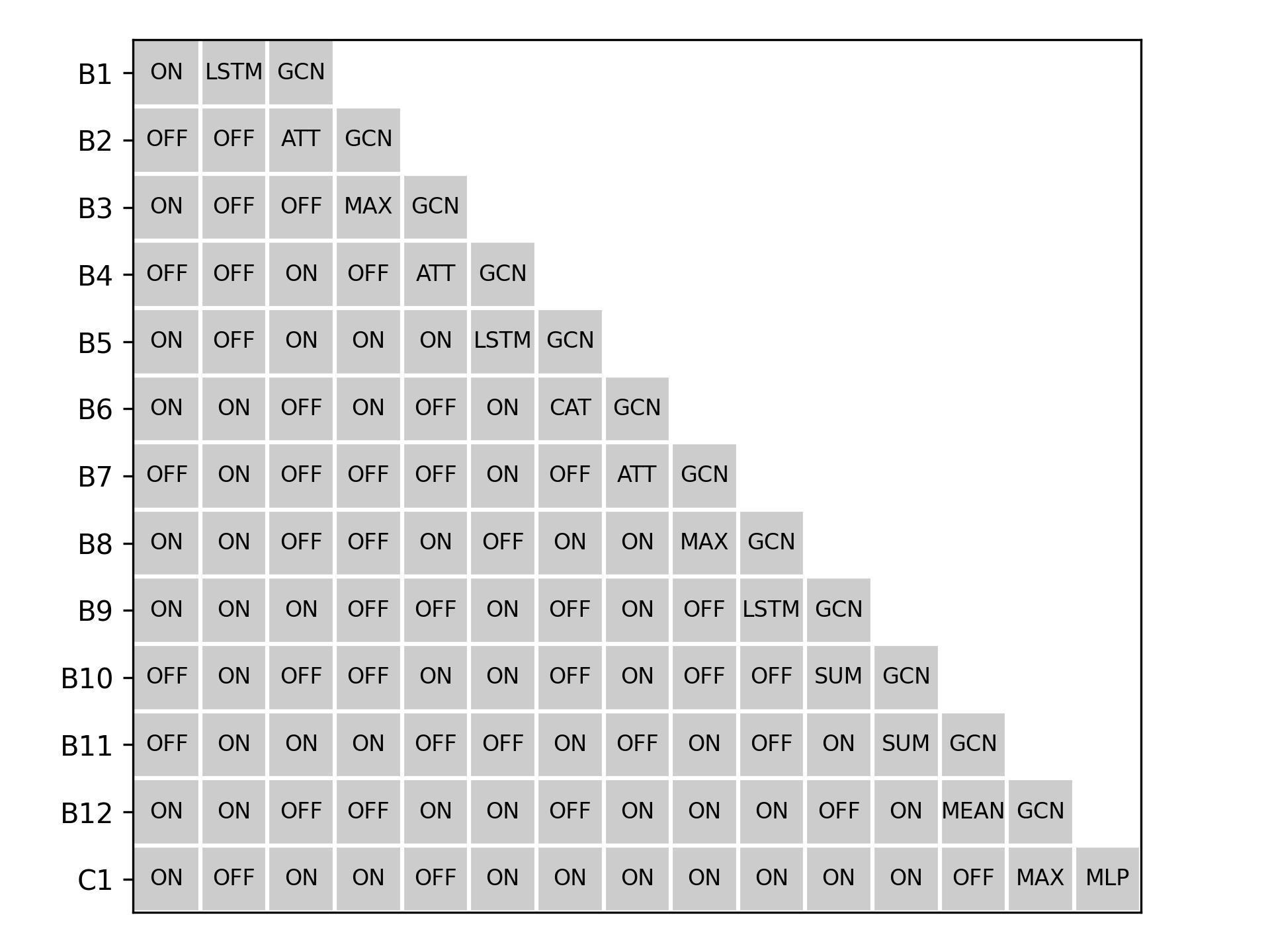}
	\includegraphics[width=0.45\linewidth]{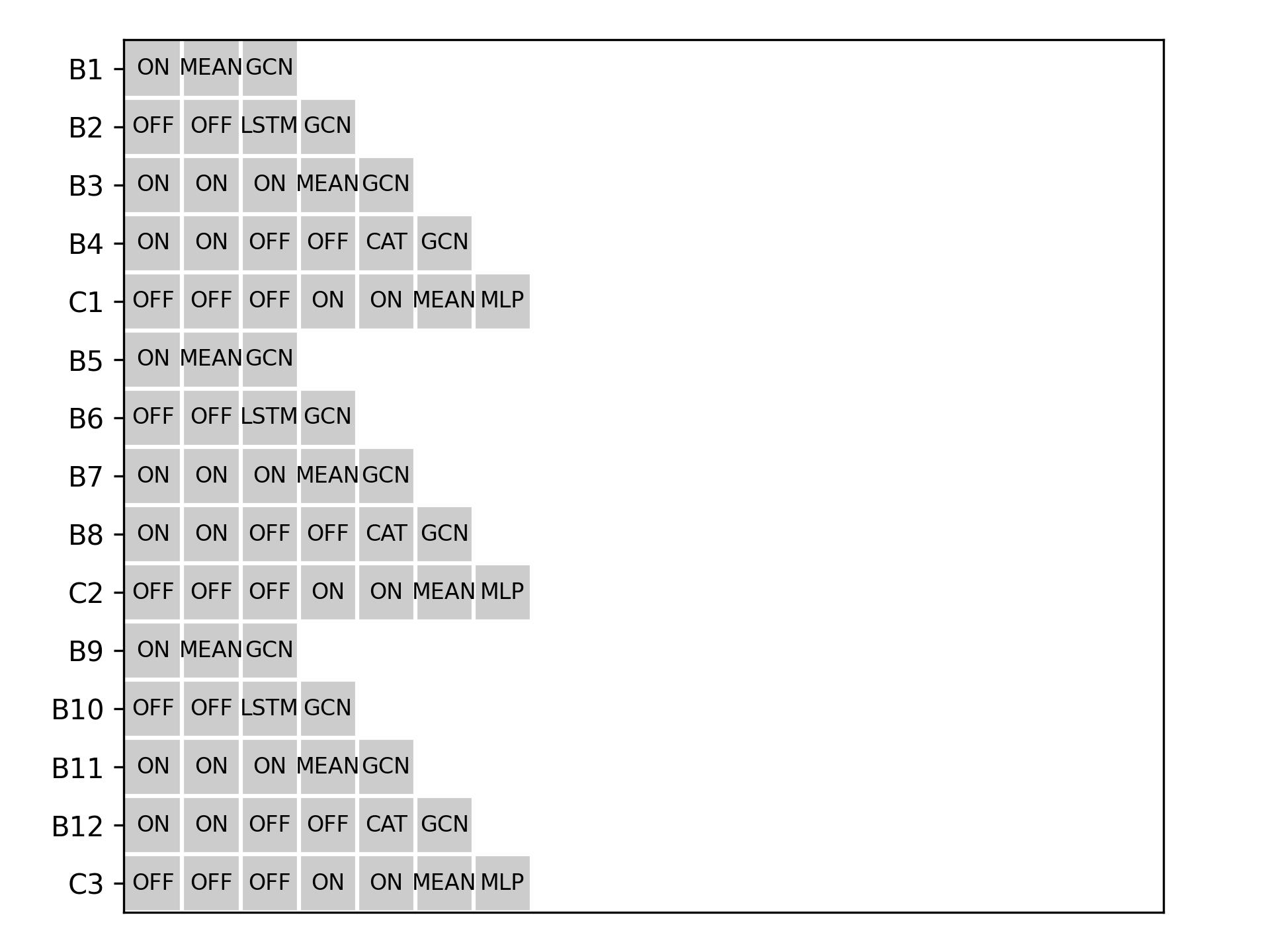}
	\includegraphics[width=0.45\linewidth]{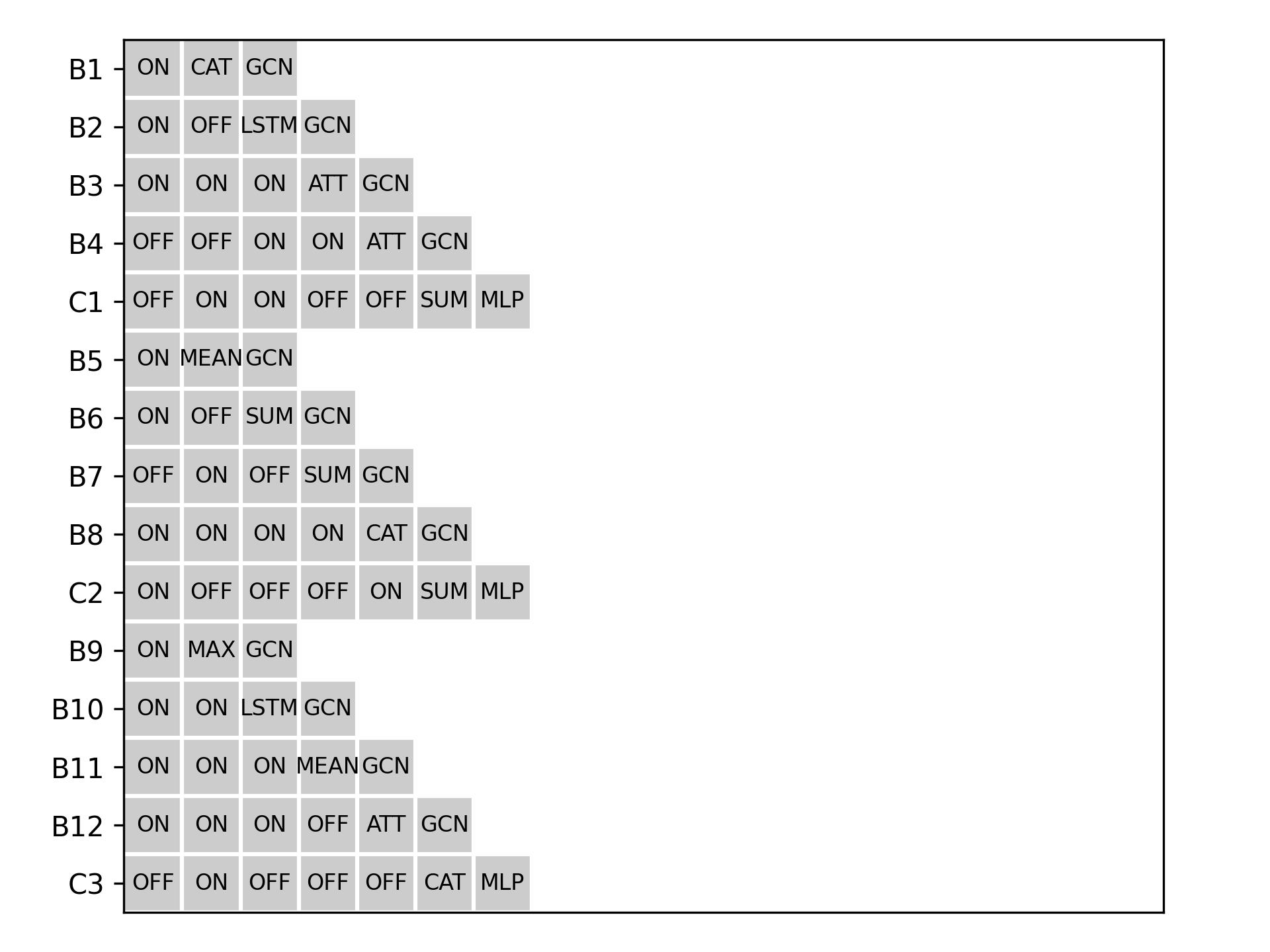}
	\caption{The searched architectures on IMDB-BINARY dataset. }
	\label{fig-searched-arch-imdbb}
\end{figure*}

The the searched GNNs are provided in Fig.~\ref{fig-searched-arch-nci1}-\ref{fig-searched-arch-imdbb}.
\label{sec-appendix-searched}
%

\subsection{Ablation Study}
\label{sec-appendix-ablation}

\subsubsection{The influence of different aggregation operations.}
\begin{figure}[ht]
	\centering
	\vspace{-10pt}
	\includegraphics[width=0.6\linewidth]{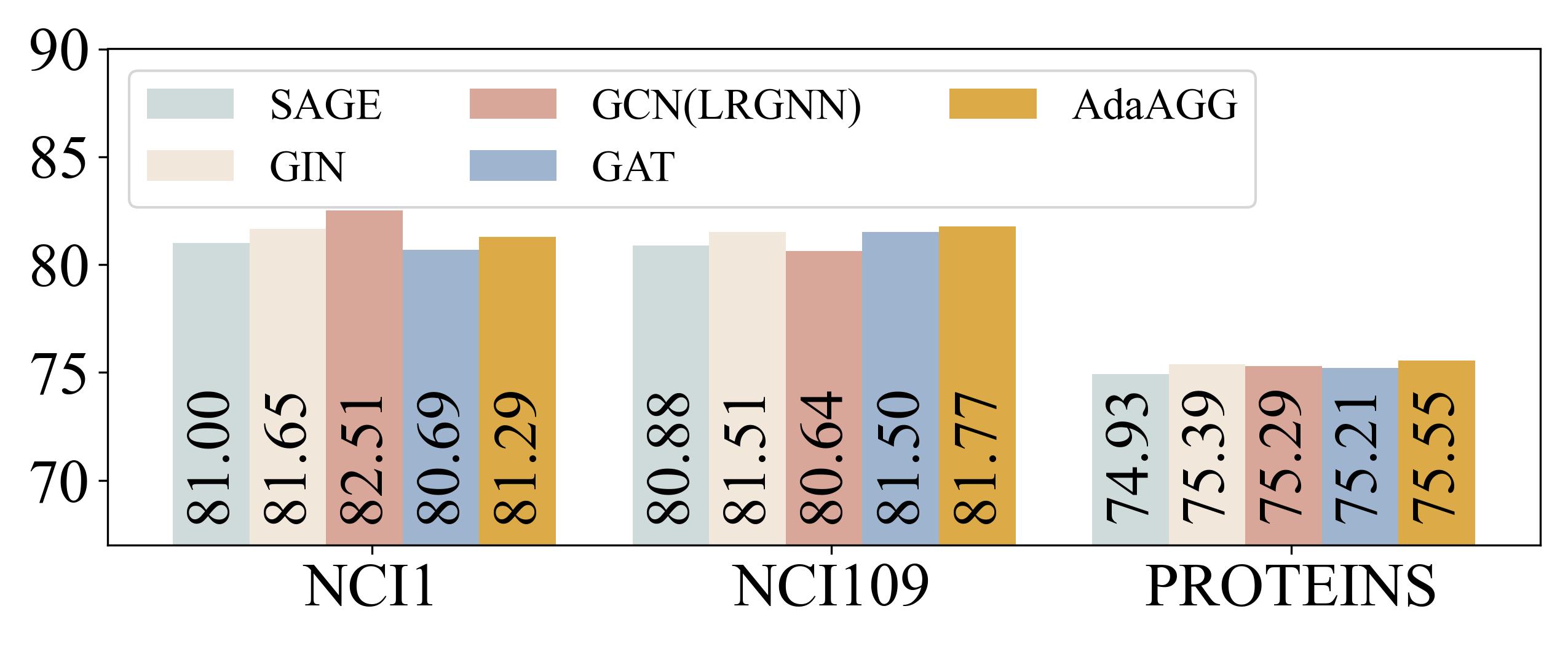}
	\vspace{-10pt}
	\caption{Evaluations on the basic aggregation operations. These methods are contructed with LRGNN-Full B12C1 variant and different aggregations. }
	
	\label{fig-ablation-agg}
\end{figure}

In this paper, we use the GCN operation to construct LRGNN. In this section, we employ different aggregation operations and evaluate their influence. As shown in Figure~\ref{fig-ablation-agg}, compared with using fixed GCN operations in our paper, using the GIN operation will achieve higher performance, which is consistent with \cite{xu2018powerful}. 
Besides, we provide one variant AdaAGG which provide the aggregation design dimension additionally, i.e., $ \{\texttt{SAGE}, \texttt{GCN}, \texttt{GIN}, \texttt{GAT}\}$. It achieves higher performance than the proposed method LRGNN which uses fixed GCN operation. It indicates that the performance can be further improved in the future by adding more effective aggregation operations in the search space.

\subsubsection{The evaluation of search algorithm}
\label{sec-ablation-algo}
\begin{figure}[ht]
	\centering
	\includegraphics[width=0.6\linewidth]{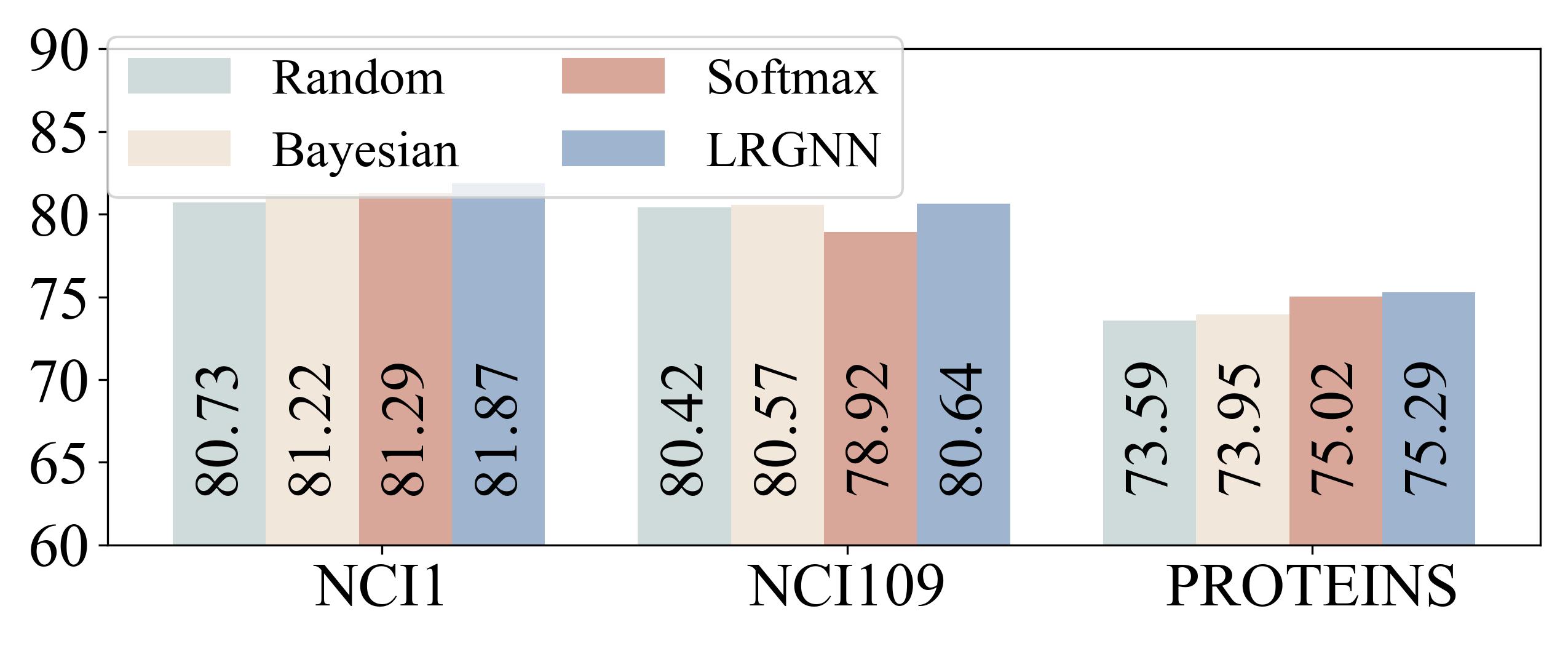}
	\vspace{-10pt}
	\caption{Evaluations on the search algorithm. We use LRGNN-Full B12C1 variant as an example. Softmax represent the differentiable algorithm with Softmax relaxation function, and our method employ the Gumbel-Softmax function instead. }
	\label{fig-ablation-algo}
\end{figure}

To make a fair comparison with LRGNN which adopts the Gumbel-Softmax function to relax the search space, we provide one baseline which uses the Softmax function as the relaxation function. We adopt the Random and Bayesian search algorithm in this section. These two methods have 100 search epochs in which one architecture is sampled from the search space and trained from scratch. We select the Top-1 architecture and tune this architecture, from which we obtained the final performance. 
As shown in Figure~\ref{fig-ablation-algo}, the Random and Bayesian baselines cannot achieve higher performance than the differentiable methods on the designed search space. Besides, compared with these two methods, the relaxation function has smaller influences and could achieve comparable performance. 
Therefore, the presence of these differentiable search algorithm is essential for good performance, and the relaxation function matter less. This function can be further updated and could achieve considerable performance in expectation.

\begin{figure*}[ht]
	\centering
	\includegraphics[width=0.45\linewidth]{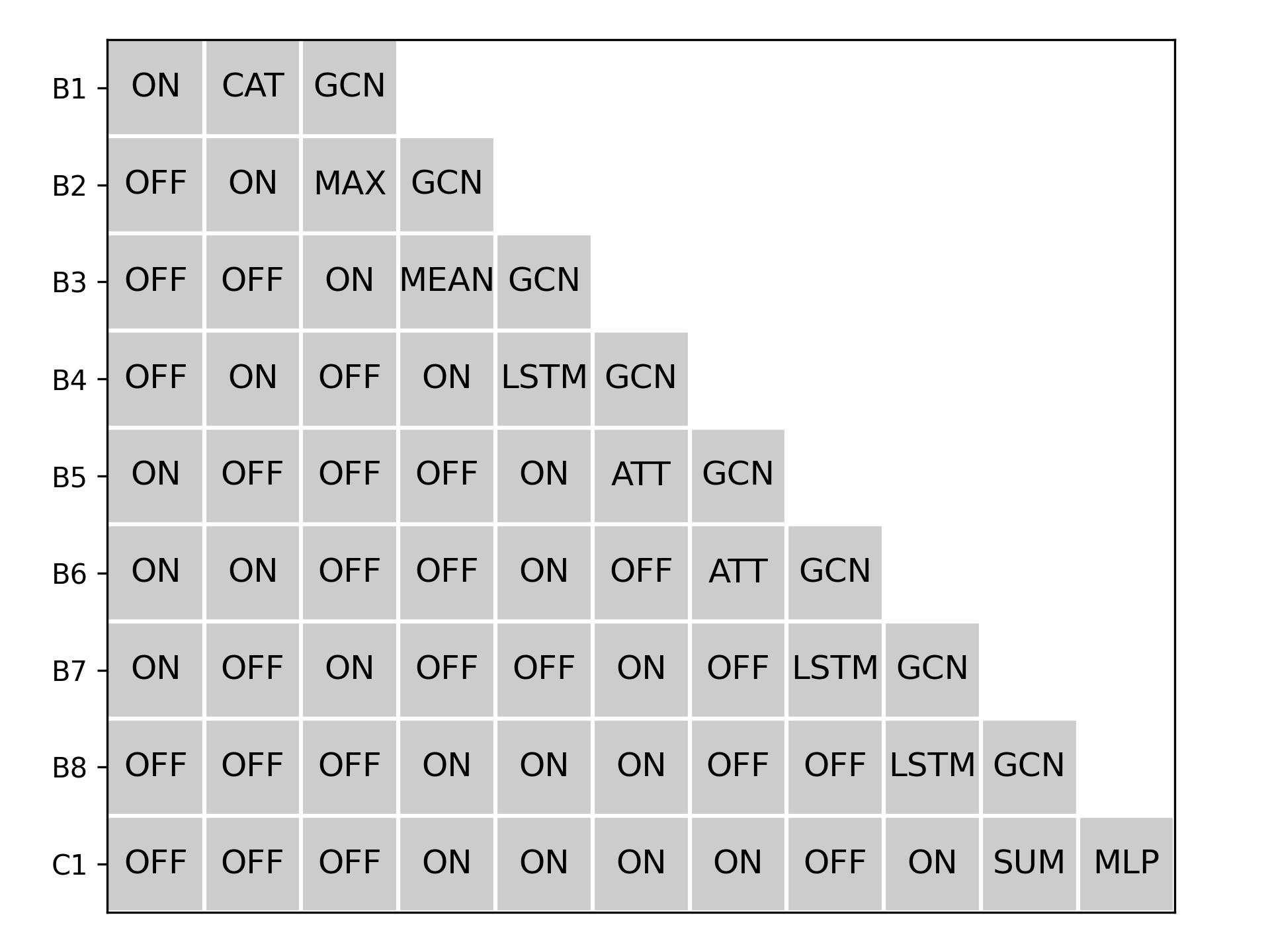}
	\includegraphics[width=0.45\linewidth]{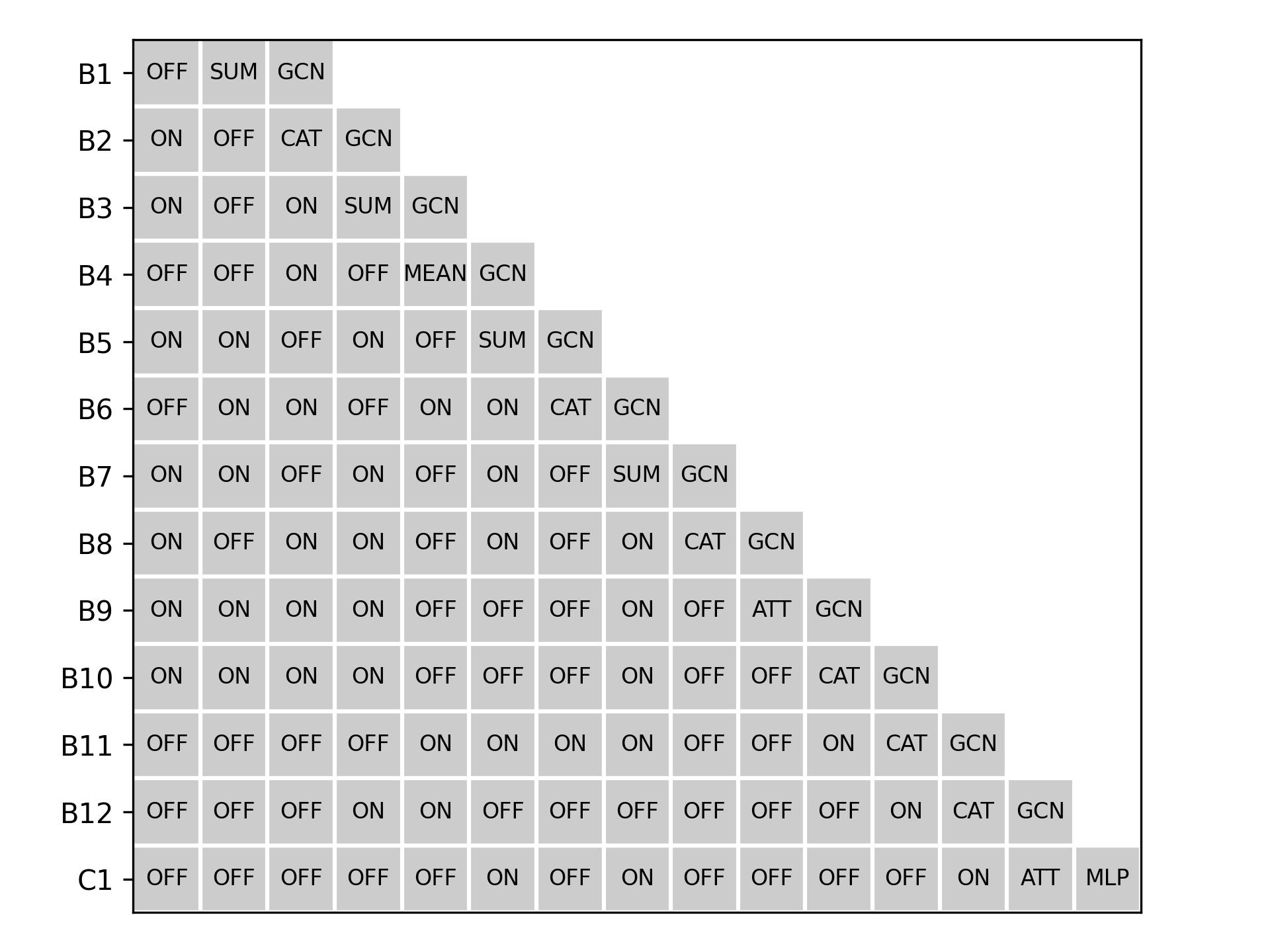}
	\includegraphics[width=0.45\linewidth]{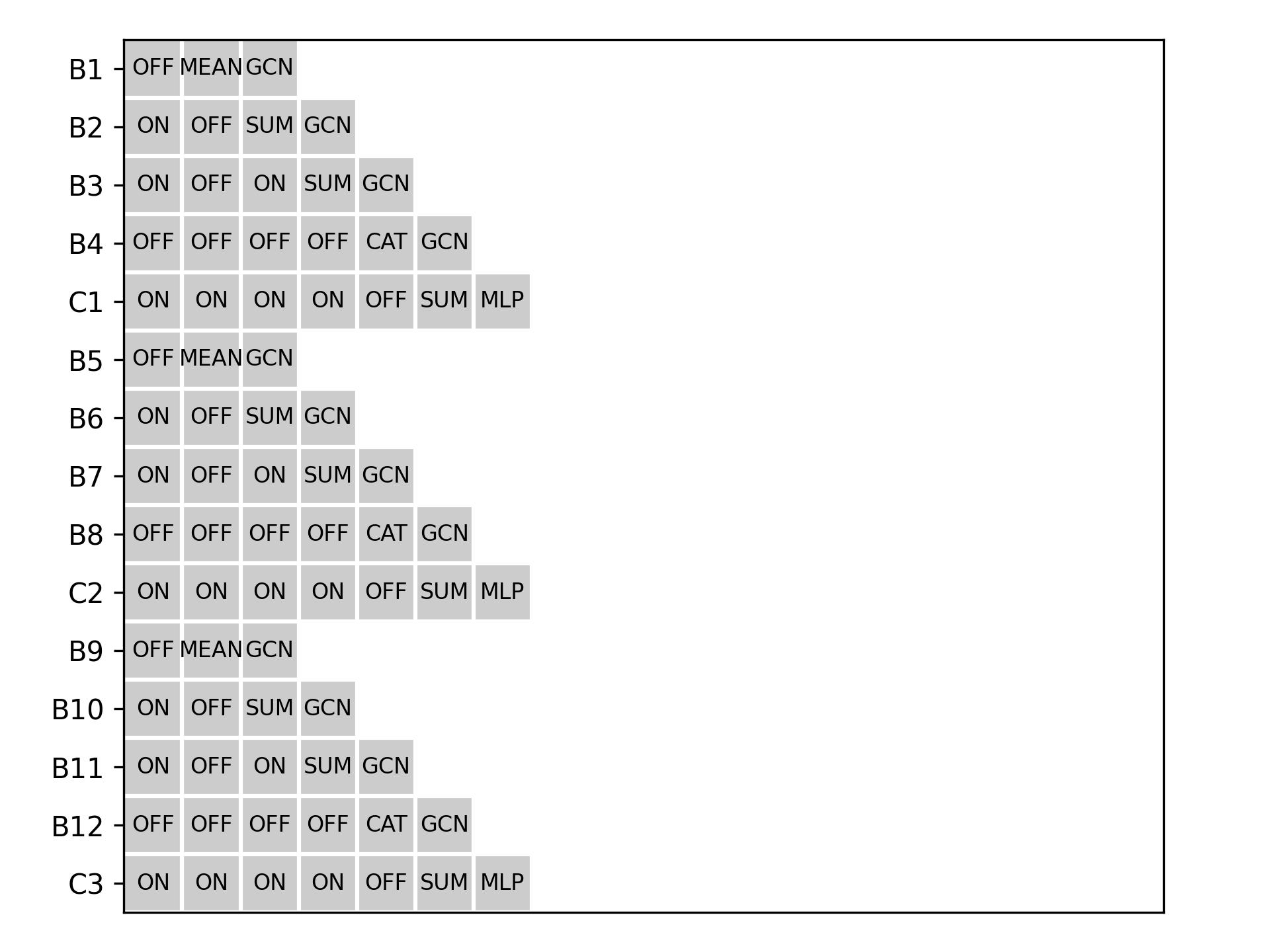}
	\includegraphics[width=0.45\linewidth]{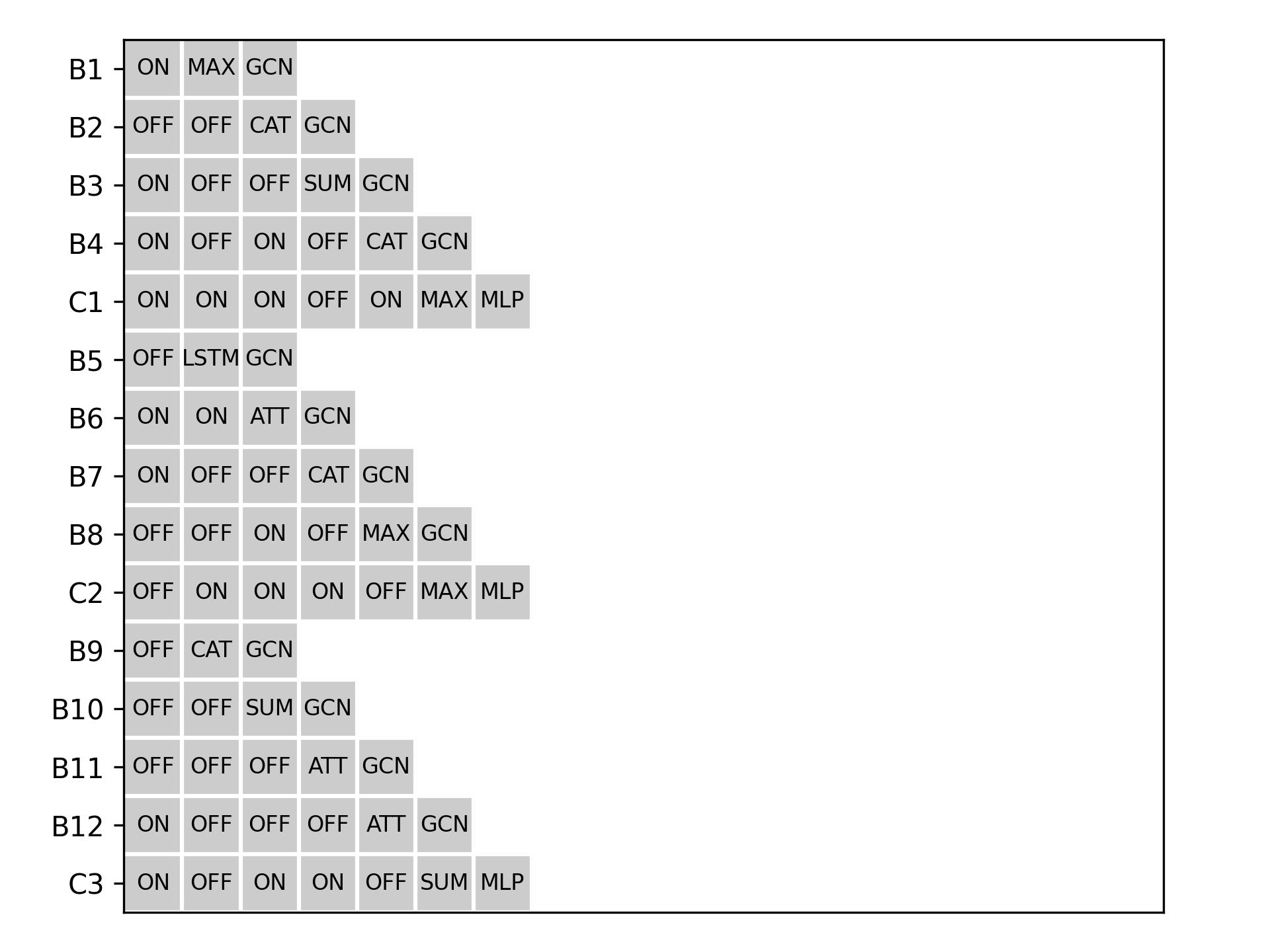}
	\caption{The searched architectures on NCI109 dataset. }
\end{figure*}
%

\begin{figure*}[ht]
	\centering
	\includegraphics[width=0.45\linewidth]{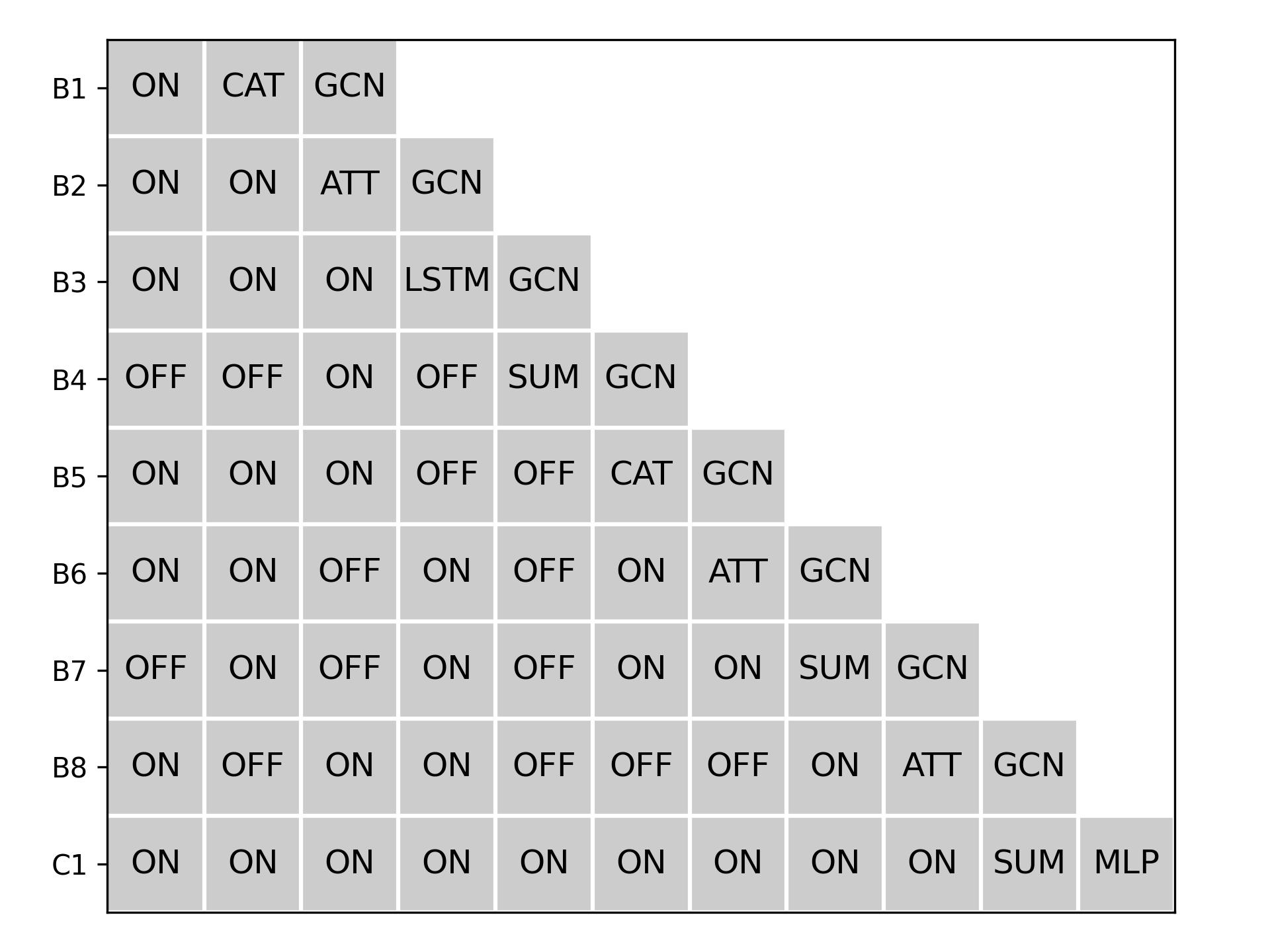}
	\includegraphics[width=0.45\linewidth]{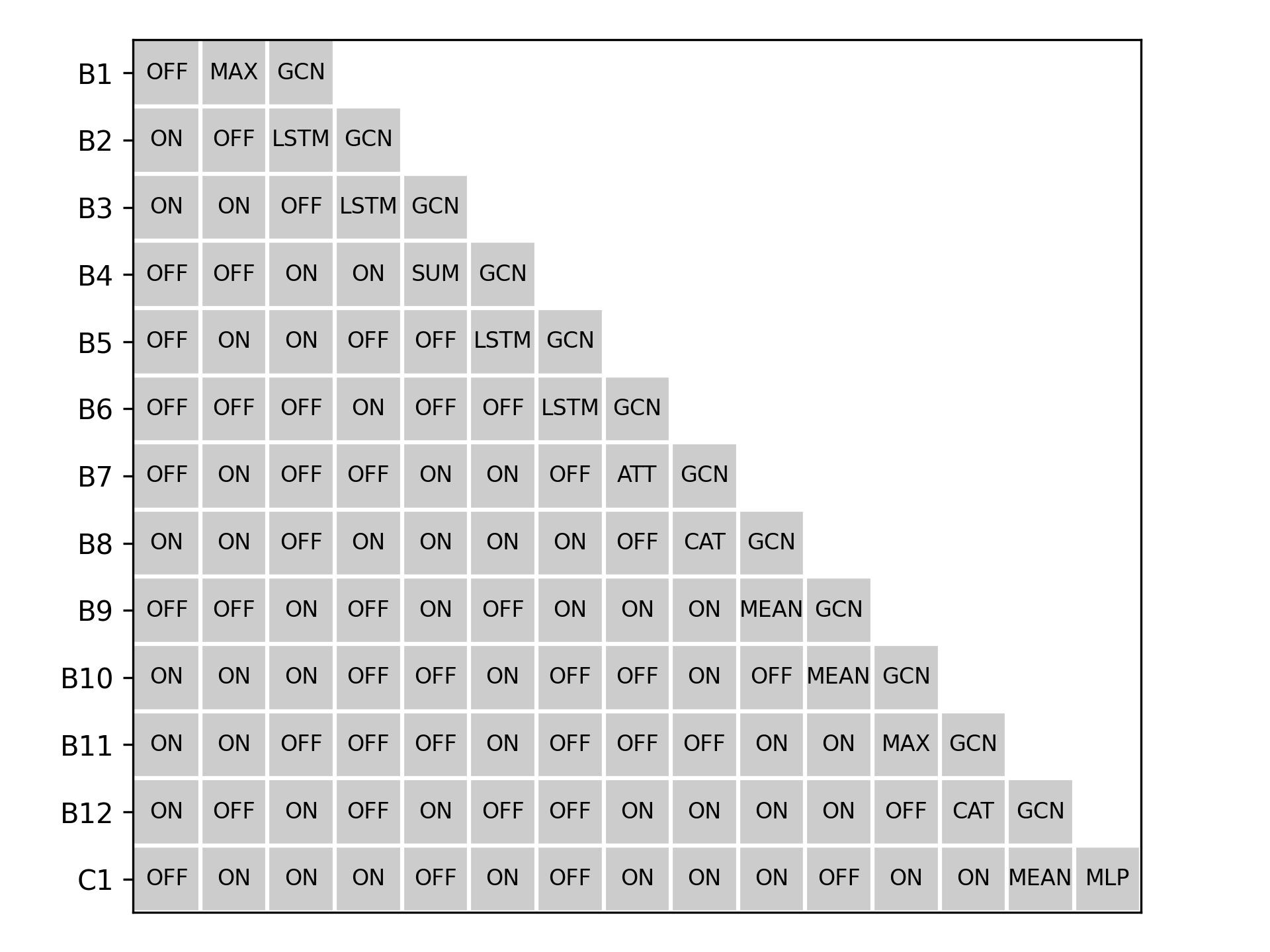}
	\includegraphics[width=0.45\linewidth]{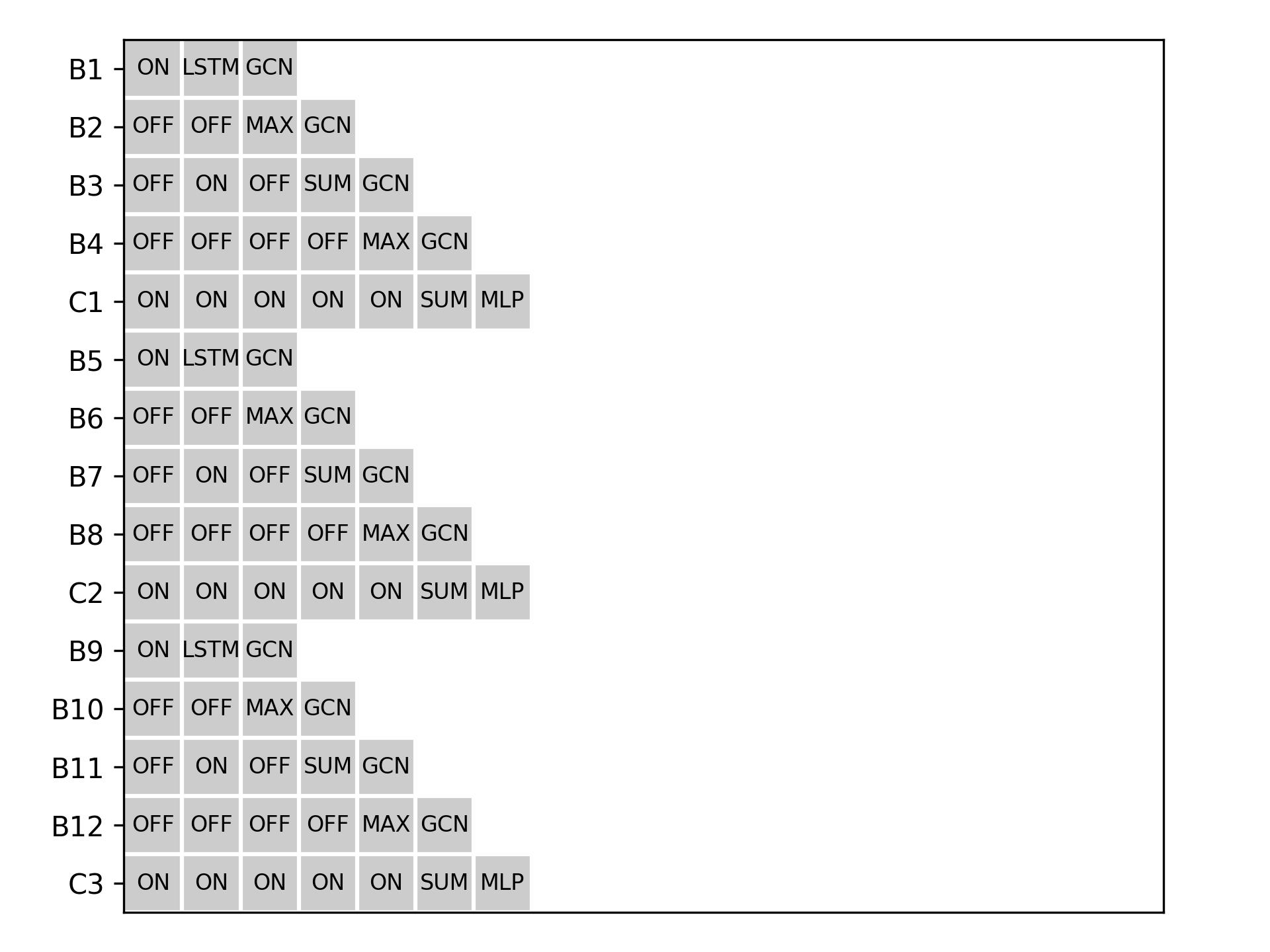}
	\includegraphics[width=0.45\linewidth]{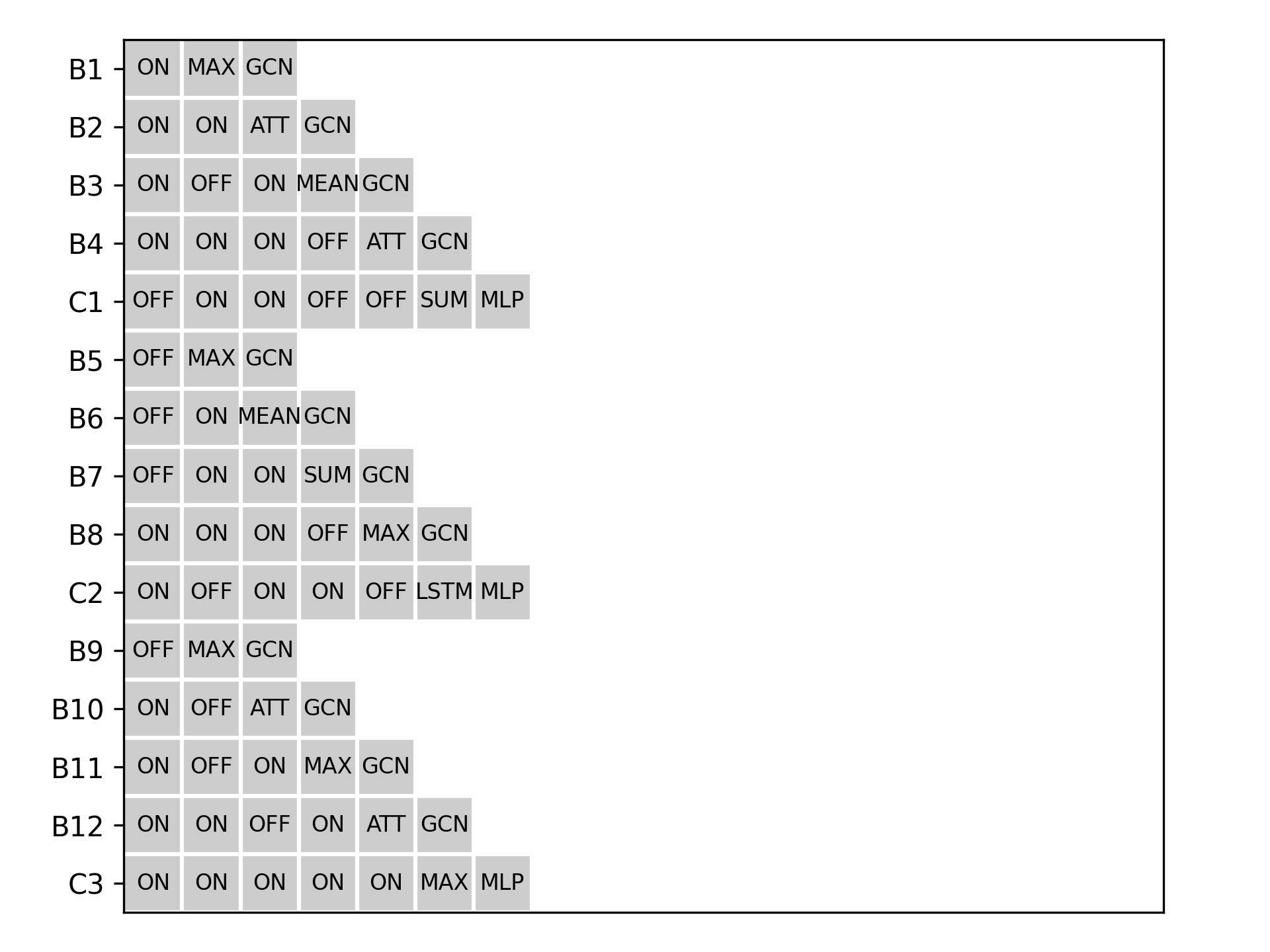}
	\caption{The searched architectures on DD dataset. }
\end{figure*}


\begin{figure*}[ht]
	\centering
	\includegraphics[width=0.45\linewidth]{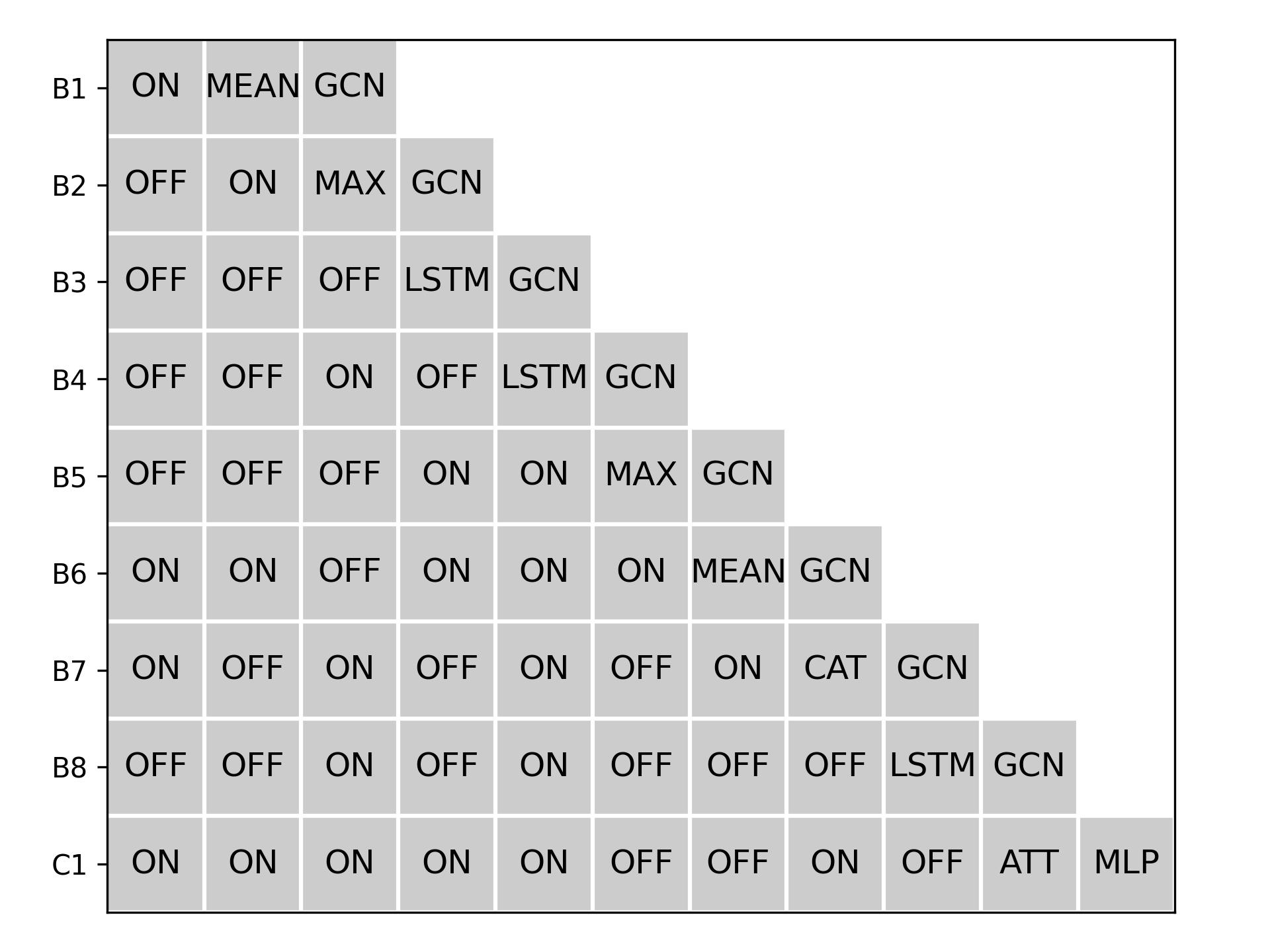}
	\includegraphics[width=0.45\linewidth]{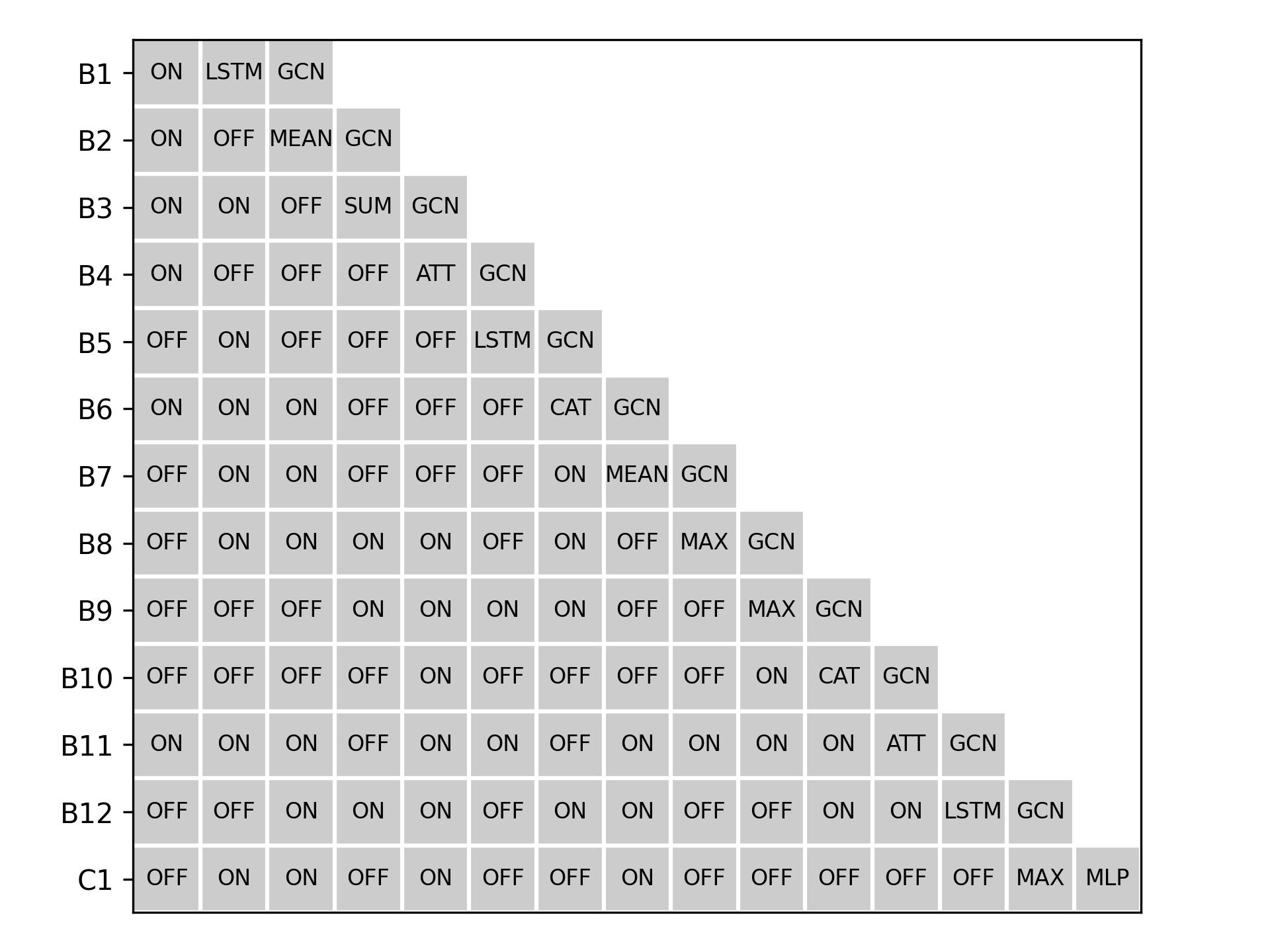}
	\includegraphics[width=0.45\linewidth]{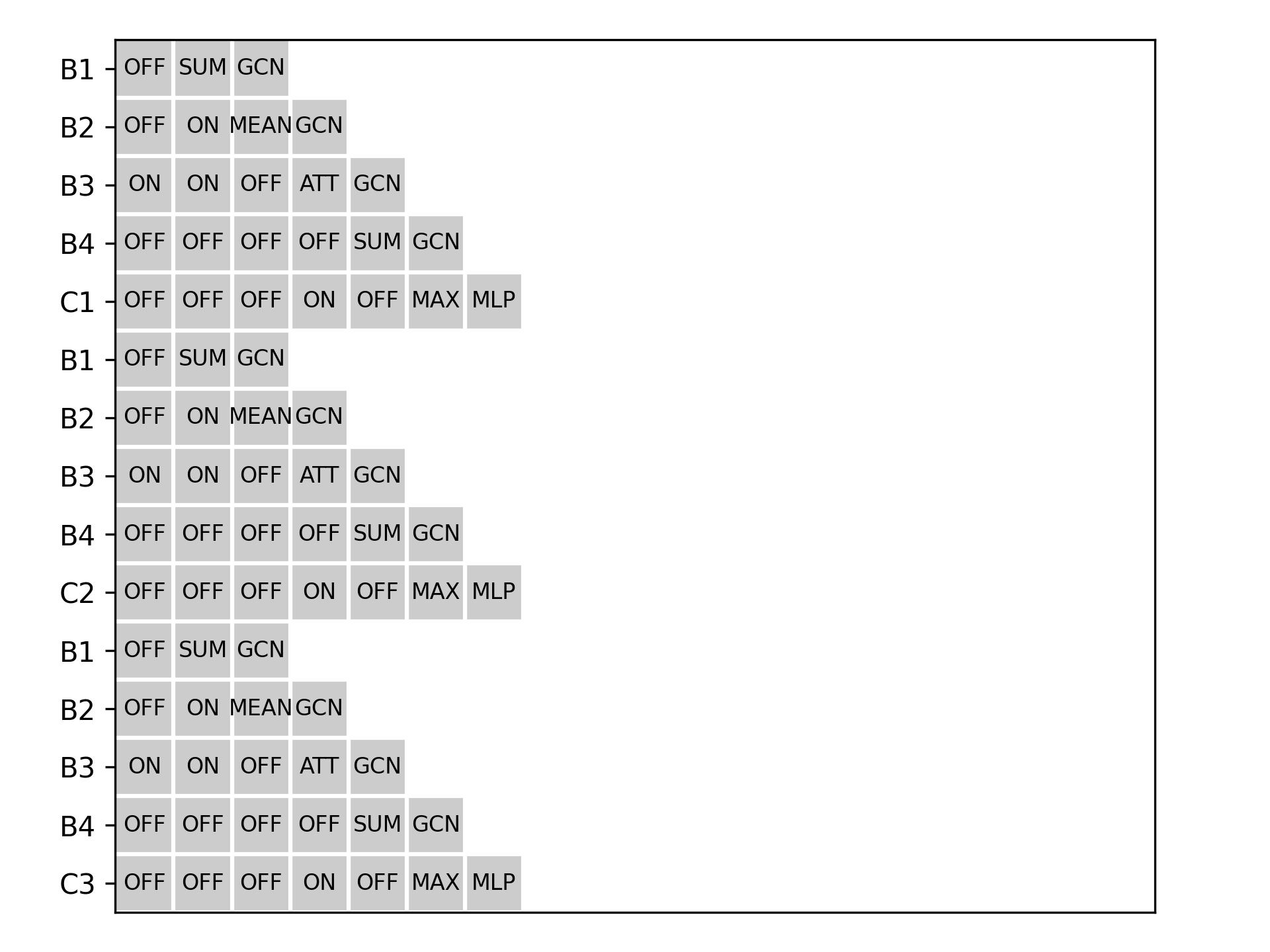}
	\includegraphics[width=0.45\linewidth]{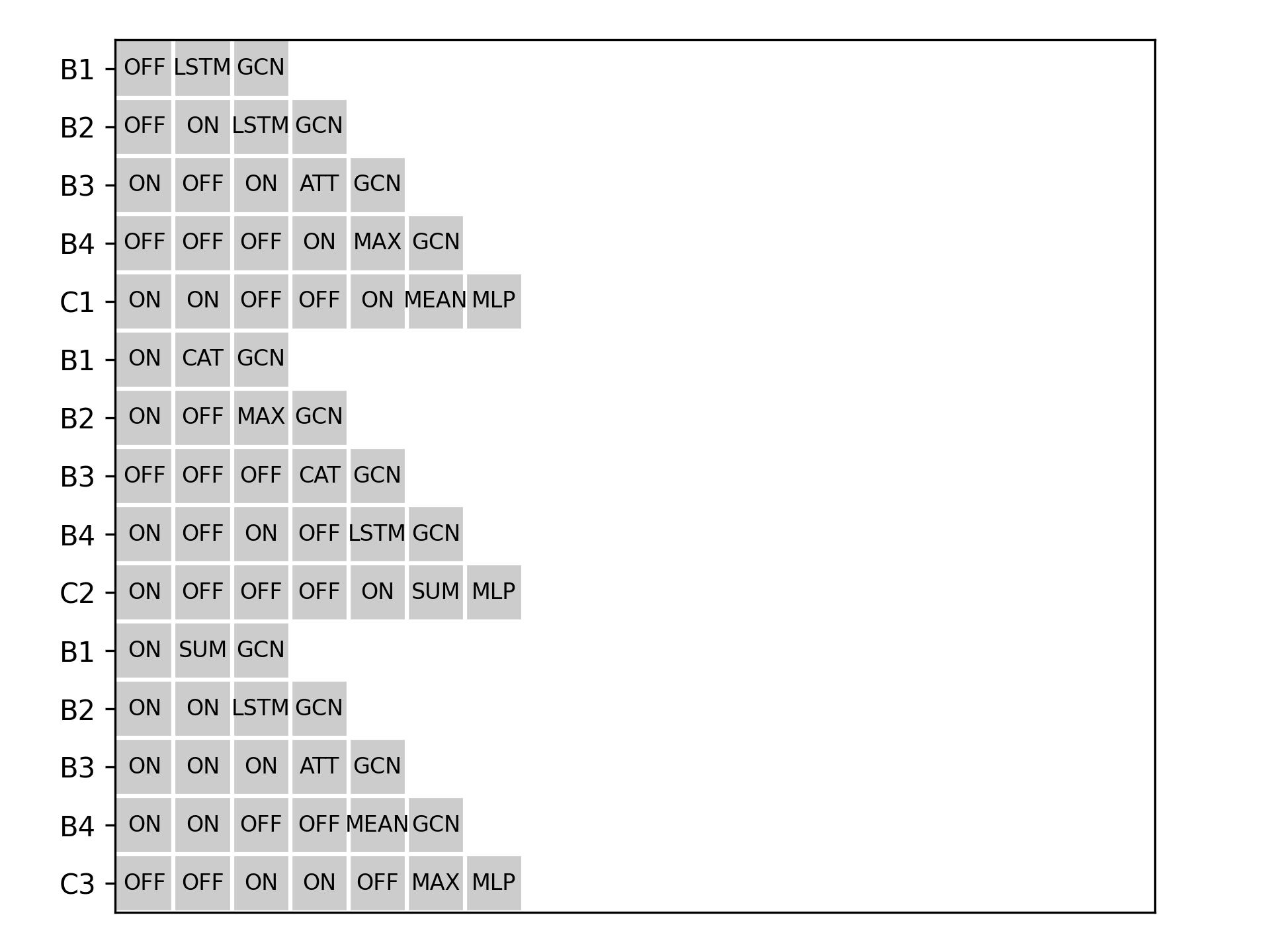}
	\caption{The searched architectures on PROTEINS dataset. }
\end{figure*}


%
%
%
%
%
%
%
%
%
%
%
%
%
%

\end{document}